\documentclass[runningheads]{llncs}

\usepackage{eccv}

\usepackage{eccvabbrv}

\usepackage{graphicx}
\usepackage{booktabs}
\usepackage{xcolor}
\usepackage[table]{xcolor}
\usepackage{pgfplots}
\pgfplotsset{compat=1.18}
\usepackage{enumitem}
\setlist[itemize]{label=\textbullet}
\usepackage{multirow}
\usepackage{array}
\usepackage{nicematrix}
\usepackage{tikz}

\usepackage{hyperref}

\usepackage[accsupp]{axessibility}  % Improves PDF readability for those with disabilities.

\usepackage{hyperref}

\usepackage{orcidlink}

\begin{document}

% \title{Anchor-Guided Attack: Parameter-Level Defenses are Insufficient Against Unauthorized Merging} 

\title{On the Vulnerability of Parameter-Level \\Defenses to Model Merging}

% Parameter-Level Defenses can be easily broken.


\author{Kuangpu Guo\inst{1,2}\orcidlink{0009-0002-7267-0428} \and
Qingyan Zheng\inst{3}\orcidlink{0009-0004-0954-1150} \and
Jian Liang\inst{2,3}\thanks{Corresponding author.}\orcidlink{0000−0003−3890−1894} \and
Yongcan Yu\inst{2}\orcidlink{0009-0000-8261-1662} \and \\
Zilei Wang\inst{1}\orcidlink{0000-0003-1822-3731} \and
Ran He\inst{2,3}\orcidlink{0000−0002−3807−991X} \and
Tieniu Tan\inst{2,3,4}\orcidlink{0000-0002-1808-2169}} 

\authorrunning{Guo. et al.}

\institute{University of Science and Technology of China \\ \and
NLPR \& MAIS, Institute of Automation, Chinese Academy of Sciences \\ \and
School of Artificial Intelligence, University of Chinese Academy of Sciences \\ \and 
Nanjing University \\
\email{gkp@mail.ustc.edu.cn, liangjian92@gmail.com}
}

\maketitle

\begin{abstract}
The training-free integration of expert models via model merging has exposed significant security risks, enabling free-riders to combine specialized models without authorization.
Recent works propose parameter-level defenses that employ linear parameter transformations to neutralize this threat.
In this paper, we systematically analyze such defenses and reveal that their protected task vectors are inherently small in magnitude.
Consequently, the protected weights remain overwhelmingly dominated by the pretrained model.
Based on this observation, we designate the pretrained model as a static reference anchor and propose the Anchor-Guided Attack (AGA) to circumvent existing safeguards. 
Specifically, AGA aligns the protected model with this anchor to recover the transformation matrix analytically.
Extensive evaluations validate that AGA consistently bypasses both individual and composite defenses under realistic defense-agnostic scenarios.
Furthermore, we provide Anchor-Repulsive Fine-tuning (ARF), a defense method to mitigate the anchor dominance leveraged by AGA.
Empirical results confirm that ARF effectively defeats the proposed attack.
Our code is available at \url{https://github.com/krumpguo/secure-merge-attack}.
\end{abstract}

\section{Introduction}
Adapting large-scale pretrained models to specialized downstream tasks via lightweight fine-tuning~\cite{wu2025llm, zhang2024scaling} has emerged as a prevailing paradigm in the deep learning community. 
This workflow is increasingly facilitated by open-source platforms, such as HuggingFace~\cite{wolf2019huggingface} and ModelScope~\cite{Modelscope}, which host a vast repository of task-specific models.
Given the proliferation of these specialized checkpoints, model merging~\cite{ilharco2022editing, guo2026stay, ding2025harmonizing} has gained prominence as a highly efficient strategy for constructing multi-task models~\cite{caruana1997multitask, zhang2018overview, zhang2021survey}.
By directly fusing the parameters of multiple fine-tuned models derived from a common pretrained backbone~\cite{limap, zhaomerging}, practitioners can effectively synthesize cross-task expertise without retraining from scratch.

\begin{figure}[t]                       
  \centering                           
  \includegraphics[width=\linewidth]{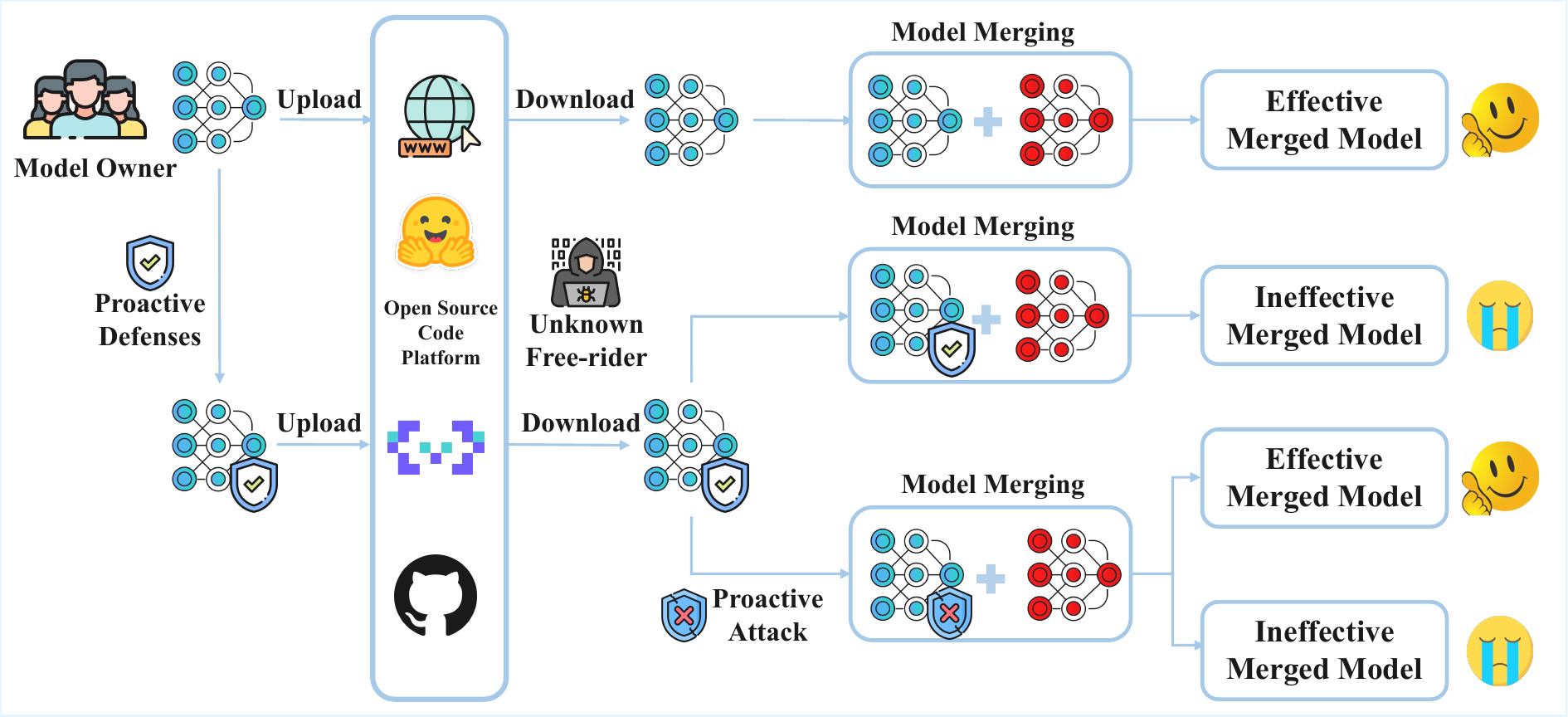}
  \caption{Illustration of proactive defense and attack in model merging.
           }
  \label{fig:pipeline}    
\end{figure}

Despite its remarkable efficiency, the inherent openness of model merging introduces severe intellectual property (IP)~\cite{cong2023have, yamabe2025mergeprint} risks. 
As illustrated in Fig.~\ref{fig:pipeline}, adversaries can effortlessly aggregate publicly available checkpoints to inherit specialized capabilities without authorization or incurring any training costs.
To prevent this illicit use, recent studies~\cite{junhao2025disrupting, wang2025model, li2025not} have proposed proactive defense mechanisms designed to intentionally destroy unauthorized merged models while preserving their standalone performance.
Specifically, by leveraging the coupled structure of model parameters (such as the natural pairing of query and key matrices),
Params~\cite{junhao2025disrupting} applies diagonal and permutation matrices to transform the MLP and attention modules. 
Similarly, MergeBarrier~\cite{li2025not} and MergeLock~\cite{wang2025model} protect the models using orthogonal and invertible matrices. 
These transformations successfully disrupt the linear connection between different fine-tuned models, leading to a severe performance drop for any unauthorized merging. 

Since existing protection methods fundamentally rely on linear transformations, the protected fine-tuned weights can be explicitly decomposed into a transformed pretrained weight $W_{pre}^P$ and a corresponding protected task vector $\tau^P$. 
A careful examination of these two components reveals a critical magnitude disparity.
As illustrated in Fig.~\ref{fig:L1_B32}, the norm of $\tau^P$ is substantially smaller than that of $W_{pre}^P$, typically by two to three orders of magnitude.
Consequently, the parameter space of the protected fine-tuned model remains heavily dominated by the protected pretrained weights.
Based on this empirical observation, we treat the pretrained model as a static reference anchor and propose the Anchor-Guided Attack (AGA) to circumvent these defenses. 
By formulating an alignment objective that minimizes the discrepancy between the protected model and this public anchor, AGA derives a recovery matrix that effectively reverses the protective transformations analytically.

To handle the diverse protection mechanisms applied across distinct architectural modules, AGA employs a dual-solver analytical framework.
For the continuous invertible or orthogonal matrices typically applied to multi-head attention modules, AGA formulates the attack as an overdetermined linear system, deriving a closed-form analytical recovery matrix via least squares regression~\cite{watson1967linear}. 
Conversely, to address the discrete permutation matrices used to shuffle the hidden neurons in MLP blocks, where continuous approximations frequently induce numerical drift, AGA casts the recovery as a linear sum assignment problem~\cite{burkard1980linear}. 
By utilizing the Hungarian algorithm~\cite{kuhn1955hungarian} with costs defined by negative cosine similarity, we achieve globally optimal discrete bipartite matching. 
Consequently, this dual-solver design ensures a mathematically rigorous and highly precise recovery of the protected architecture.

We validate the effectiveness of our attack through a comprehensive cross-modal empirical study spanning both computer vision and natural language processing. 
Extensive experiments demonstrate that in realistic, defense-agnostic scenarios, AGA consistently bypasses state-of-the-art defenses.
Specifically, upon applying our attack, the performance of the unauthorized merged model recovers up to 97\% of the results achieved by an unprotected merged model. 
Furthermore, our approach successfully compromises both standalone defense mechanisms and their sophisticated composite configurations.

To mitigate the security vulnerabilities exposed by AGA, we investigate adaptive countermeasures.
Given that the attack fundamentally exploits the marginal norm of task vectors relative to their pretrained anchors, a robust defense requires proactively amplifying this norm during the fine-tuning phase.
Therefore, we introduce Anchor-Repulsive Fine-tuning (ARF), a defensive strategy that employs a distance-based repulsive term specifically within the attention modules to eliminate the magnitude disparity leveraged by AGA.
Experimental results demonstrated that ARF effectively prevents unauthorized merging while maintaining the standalone model's utility with negligible performance degradation.
Our contributions can be summarized as follows:
\begin{itemize}
\item We perform a systematic analysis of parameter-level defenses and reveal that protected task vectors are orders of magnitude smaller than pretrained weights.
Consequently, the parameter space of protected models remains heavily dominated by the pretrained backbone.
\item We propose AGA, the first universal attack framework capable of circumventing state-of-the-art parameter-level protections.
By employing a specialized dual-solver design, AGA analytically recovers protective transformations, consistently bypassing both individual and composite defenses in realistic, defense-agnostic scenarios.
\item We introduce ARF, an adaptive defense strategy designed to neutralize the threat posed by AGA.
By applying a localized repulsive force within the attention modules, ARF effectively prevents unauthorized merging while preserving the model’s standalone utility.
\end{itemize}

\section{Preliminary}
\subsection{Model Merging and Task Arithmetic}
Let $\theta_{pre}$ denote the parameters of the pre-trained model, and $\theta_{i}$ the parameters of a model fine-tuned on task $D_{i}$.
For $n$ task-specific models \{ $\theta_{1}, ..., \theta_{n}$ \} derived from the same pretrained model $\theta_{pre}$, model merging can be formulated as $\theta_{m}$ = $\mathcal{M}$($\theta_{pre}, \theta_{1}, ..., \theta_{n}$), where  $\mathcal{M}$ represents a specific parameter-level fusion strategy.
A prevalent model merging method is Task Arithmetic~\cite{ilharco2022editing}, which operates on task vectors defined as follows:
\begin{equation}
    \tau_i = \theta_{i} - \theta_{pre}.
\end{equation}

To merge $n$ expert models $\{\theta_1, \theta_2, ..., \theta_n\}$, the merged model $\theta_{m}$ is constructed by applying a linear combination of their respective task vectors $\{\tau_i\}_{i=1}^n$ to the pre-trained model:
\begin{equation}
    \theta_{m} = \theta_{pre} + \lambda \sum_{i=1}^n  \tau_i,
\end{equation}
where $\lambda$ is a scaling factor.

\subsection{Problem Formulation}
\textbf{Attack Scenario.} 
In this paper, we consider two parties: the defender and the adversary. 
The defender fine-tunes a pretrained model on specialized data and releases it to the broader community to demonstrate its capabilities.
The adversary acquires the open-source model and merges it with other models they control, all of which are fine-tuned from the same pretrained model.

\noindent \textbf{Defender’s Capability.}
The defender has full control over the model's lifecycle. 
This includes both the training and post-training stages to render the model unmergeable while preserving its original task utility. 

\noindent \textbf{Adversary’s Capability.}
The adversary has access to the public pre-trained model $\theta_{pre}$ and the protected proprietary model $\theta^p_{ft}$. 
Importantly, the adversary is agnostic to the protection state, meaning they do not know whether a specific model has been secured or which defense mechanism is deployed.

\noindent \textbf{Problem Setup.} 
Let $ Perf (\theta; \mathcal{D})$ denote the performance of a model with parameters $\theta$ on task $\mathcal{D}$.
Let $\theta_{m}$ be the result of merging an unprotected model $\theta_A$ with another model $\theta_B$, and ${\theta}_{m}^{p}$ be the result of merging a protected version ${\theta}_A^{p}$ with $\theta_B$.
A defense is successful if the unauthorized merging of the protected model results in a significant performance collapse on task $\mathcal{D_A}$, 
\begin{equation}
    Perf({\theta}_{m}^{p}; \mathcal{D_A}) \ll Perf(\theta_{m}; \mathcal{D_A}),
\end{equation}
while the individual performance of ${\theta}^{p}_A$ remains preserved.
An attack is considered successful if it can recover a model $\theta^{a}_{A}$ from ${\theta}_A^{p}$ such that the resulting merged model $\theta_{m}^{a}$ restores the original performance as follows:
\begin{equation}
    Perf ({\theta}_{m}^{a}; \mathcal{D_{A,B}}) \approx Perf(\theta_{m}; \mathcal{D_{A,B}}).
\end{equation}

\subsection{Existing Defense Mechanisms}
\label{Existing Defense Mechanisms}
To safeguard proprietary expert models, several state-of-the-art defenses have been proposed. These methods generally apply secret parameter transformations to disrupt the linear arithmetic required for model merging while maintaining single-task utility.

\noindent \textbf{Params and Params-D}. 
The Params~\cite{junhao2025disrupting} defense introduces secret linear transformations into the weight space to disrupt unauthorized merging while preserving single-task utility. 
For a two-layer MLP defined as:
\begin{equation}
    MLP(X) = W_2 \sigma(W_1 X + b_1) + b_2,
\end{equation}
the protected weights ($W_{1}^p, W_{2}^p, b_1^p$) are formulated as:
\begin{equation}
    W_{1}^p = P W_{1}, \quad W_{2}^p = W_{2}P^T, \quad b_1^p = P b_1,
\end{equation}
where $P$ is a secret permutation matrix with exactly one "1" in each row and column and "0"s elsewhere.
For the multi-head attention module, the standard operation with its corresponding output projection is defined as:
\begin{equation}
    Attention = \text{softmax}\left(\frac{W_QW_K^T}{\sqrt{d}}\right)W_VW_O.
\end{equation}
Params~\cite{junhao2025disrupting} safeguards the weights of each attention head by applying two independent diagonal matrices, $A$ and $B$, as follows:
\begin{equation}
    W_Q^p = W_QA, \quad W_K^p = W_KA^{-1}, \quad W_V^p = W_VB, \quad W_O^p = B^{-1}W_O.
\end{equation}
An advanced variant, Params-D~\cite{junhao2025disrupting}, further complicates unauthorized extraction by applying random dropout to the transformed weights.

\noindent \textbf{MergeBarrier}.
Building upon similar parameter-space transformation principles, this defense~\cite{li2025not} protects the attention mechanism through orthogonal matrix transformations, while securing the MLP layers by approximating the activation functions via Taylor expansion to prevent unwanted merging.

\noindent \textbf{MergeLock}.
This method~\cite{wang2025model} secures the attention modules by applying secret invertible matrices to disrupt weight alignment.
The details of these defense methods and more related work can be found in the supplementary material.

\begin{figure}[t]
\captionsetup[subfigure]{font=small}
  \centering
  \setlength{\abovecaptionskip}{0.2cm} 
  
  \begin{subfigure}[c]{0.48\textwidth}
    \centering
    \includegraphics[width=\linewidth]{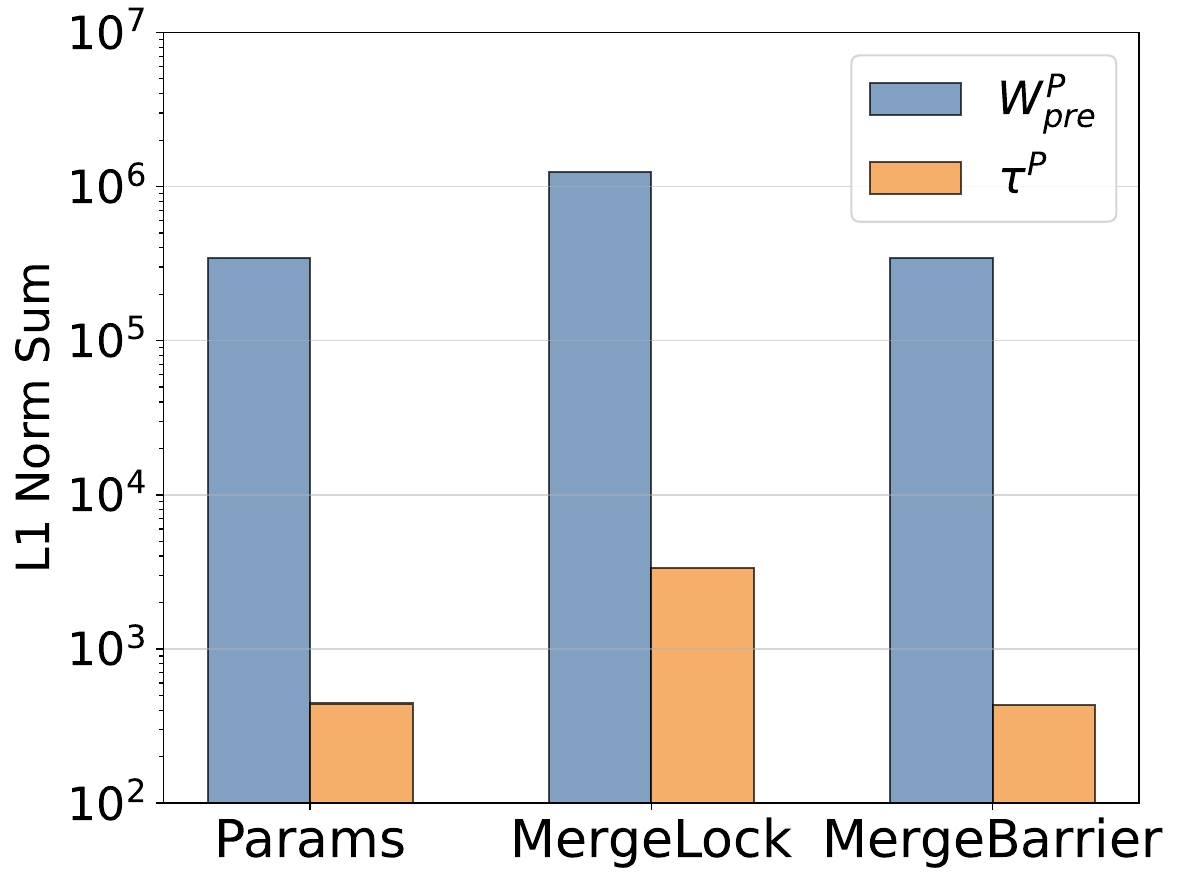}
    \caption{Comparison of Frobenius norm.}
    \label{fig:L1_B32}
  \end{subfigure}
  \hfill 
  \begin{subfigure}[c]{0.48\textwidth}
    \centering
    \includegraphics[width=\linewidth]{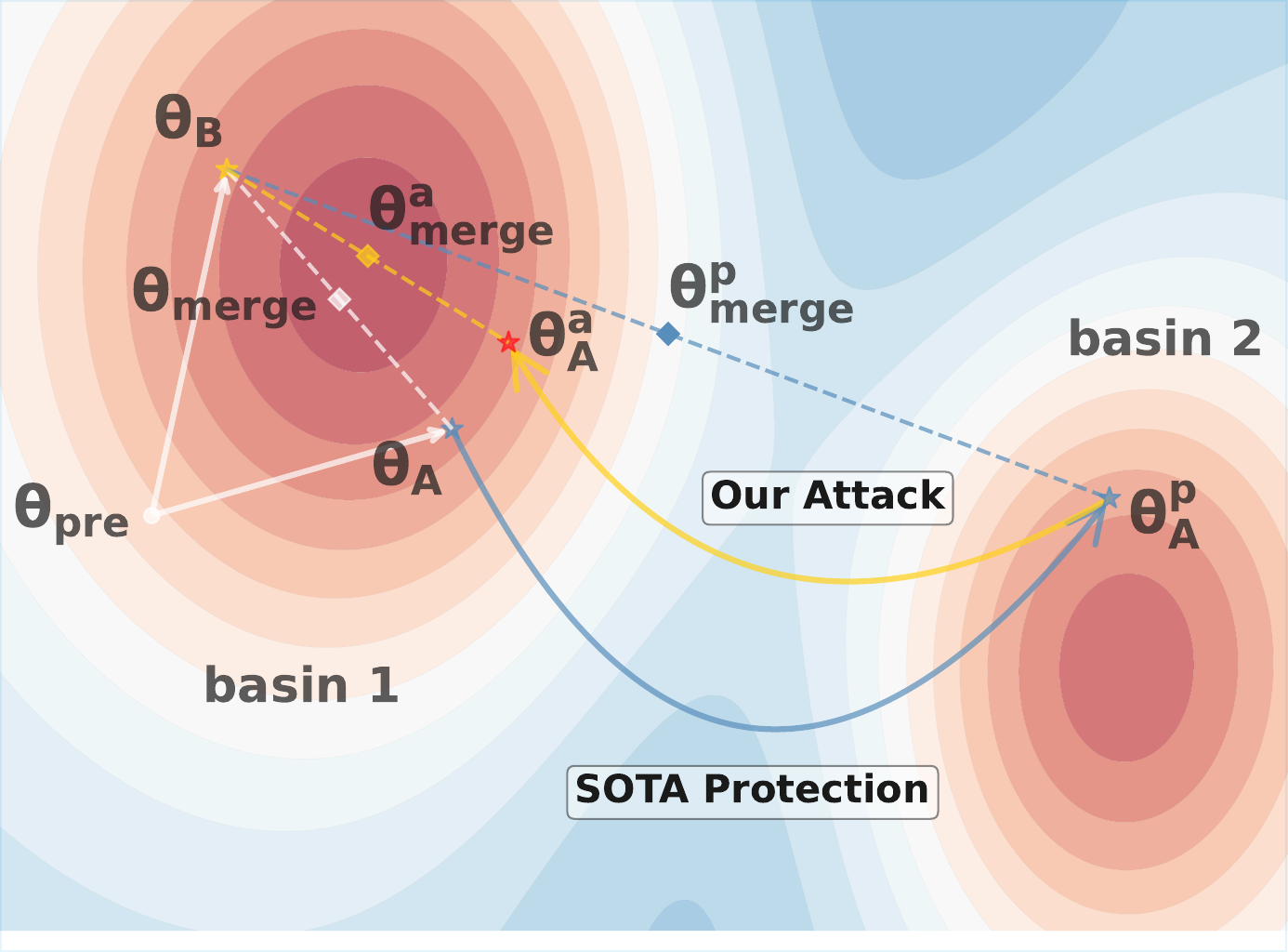}
    \caption{Loss landscape.}
    \label{fig:attack_loss}
  \end{subfigure}

  \caption{Analysis of parameter magnitude and optimization landscape.  
  (a) Comparison of the Frobenius norm between the protected pretrained weights ($W_{pre}^P$) and the protected task vector ($\tau^P$) on the ViT-B/32 model finetuend on Cars. 
  (b) Loss landscape illustration of our attack against the state-of-the-art (SOTA) protections.}
  \label{fig:combined_L1_loss}
\end{figure}

\section{Proposed Attack Method}
\label{Attack Method}
In the context of model merging, the fine-tuned weights can be formulated as $W_{ft} = W_{pre} + \tau$.
When a defense mechanism applies a secret linear transformation $P$ (e.g., an invertible or permutation matrix~\cite{junhao2025disrupting, wang2025model, li2025not}) to protect the model, the transformation is distributed across both components:
\begin{equation}
    W^p = W_{ft} P = (W_{pre} + \tau) P = W_{pre} P + \tau P.
\end{equation}
Our methodology is driven by a critical empirical observation regarding this distribution: the magnitude of the transformed task vector $\tau P$ is substantially smaller than that of the transformed anchor $W_{pre} P$, typically by two to three orders of magnitude (as illustrated in Fig.~\ref{fig:L1_B32}).
Consequently, the parameter space of the protected model is overwhelmingly dominated by the $W_{pre} P$ term, rendering the contribution of $\tau P$ virtually negligible in comparison.

This magnitude disparity directly motivates our Anchor-Guided Attack (AGA). By leveraging the approximation $W^p \approx W_{pre} P$, we optimize a recovery matrix $T$ to realign the protected weights with the public pretrained anchor, targeting the objective $W^p T = W_{pre}$. 
Because the dominant anchor term overwhelmingly dictates the optimization landscape, this alignment mathematically forces $P T \approx I$, effectively converging to the exact inverse $T \approx P^{-1}$. 
Therefore, as shown in Fig.~\ref{fig:attack_loss}, AGA seamlessly bypasses the defense and extracts the original fine-tuned weights ($W^a$) without requiring any knowledge of the defense specifics:
\begin{equation}
    W^a = W^p T \approx (W_{pre} P + \tau P) P^{-1} = W_{ft}.
\end{equation}
Building upon this core insight, AGA deploys two tailored analytical solvers: one for continuous transformations within the attention modules, and another for discrete permutations within the MLP modules.

\subsection{Continuous Attack for Attention via Least Squares}
Previous defenses typically protect the multi-head attention~\cite{vaswani2017attention} modules by applying continuous invertible, orthogonal, or diagonal matrices. 
Given our alignment objective $W^p T = W_{pre}$, recovering the continuous matrix $T$ is mathematically equivalent to fitting an overdetermined linear system. 
While defenders often apply coupled transformations across specific modules (e.g., structurally paired Query and Key), a realistic adversary remains strictly agnostic to these secret defense configurations.
To circumvent this without losing generalizability, AGA strategically decouples the estimation process, solving an independent linear system for each projection matrix.

Formally, for any protected attention matrix  $W^p \in \{W_Q^p, W_K^p, W_V^p, W_O^p\}$ and its corresponding pretrained anchor $W_{pre}$, we formulate a least squares~\cite{watson1967linear} objective to solve for the universal transformation $T$ empirically:
\begin{equation}
\min_{T} \| W^p T - W_{pre} \|_F^2,
\end{equation}
where $\| \cdot \|_F$ denotes the Frobenius norm.
Setting the derivative to zero yields the closed-form analytical inverse:
\begin{equation}
T^* = ((W^p)^T W^p)^{-1} (W^p)^T W_{pre}.
\end{equation}
Applying this closed-form solution independently to each module yields a distinct optimal recovery matrix ($T_Q^*, T_K^*, T_V^*$, and $T_O^*$). 
The fully recovered attention weights are thus explicitly formulated as:
\begin{equation}
W_Q^{a} = W_Q^p T_Q^*, \quad W_K^{a} = W_K^p T_K^*, \quad W_V^{a} = W_V^p T_V^*, \quad W_O^{a} = W_O^p T_O^*.
\end{equation}
This parallel extraction effectively circumvents all deployed continuous defenses, restoring the attention mechanism's original parameter space.

\subsection{Discrete Attack for MLP via Linear Sum Assignment}
Due to the presence of non-linear activation functions in MLP layers, defenses typically employ discrete permutation matrices to shuffle hidden neurons, ensuring that standalone model performance remains invariant.
While continuous least-squares regression can theoretically approximate this inverse permutation, it fails to enforce strict discrete constraints.
This inevitably introduces numerical drift and accumulation errors across layers.

To overcome this, AGA casts the permutation recovery as a linear sum assignment problem~\cite{burkard1980linear}.
Let $W_1^p$ be the protected first-layer MLP weight with shuffled rows, and $W_{pre}$ be the corresponding pretrained weights.
We construct a pairwise cost matrix $C$ measuring the negative cosine similarity between their respective row vectors:
\begin{equation}
    C_{i, j} = - \frac{W^p_{i} \cdot W^{pre}_{j}}{\| W^p_{i} \|_2 \| W^{pre}_{j} \|_2},
\end{equation}
where $W^p_{i}$ and $W^{pre}_{j}$ denote the $i$-th and $j$-th row vectors of $W^p_1$ and $W_{pre}$, respectively.

With the cost matrix $C$ constructed, we employ the Hungarian algorithm~\cite{kuhn1955hungarian} to explicitly find the globally optimal discrete mapping as follows:
\begin{equation}
    T^* = \arg\min_{T} \sum_{i} \sum_{j} C_{i,j} T_{i,j},
\end{equation}
where $T$ is constrained to be a valid permutation matrix. This yields the exact recovery matrix $T^*$ that minimizes the total matching cost, perfectly realigning the shuffled rows:
\begin{equation}
    W^{a}_1 = T^* W^p_1.
\end{equation}
To preserve the equivalent input-output mapping of the entire MLP block, AGA must concurrently restore the subsequent layers.
Since $T^*$ is a permutation matrix, its inverse is simply its transpose $(T^*)^T$. 
Thus, AGA strictly applies this inverse mapping to the columns of the subsequent projection layer and the intermediate biases.
\begin{equation}
W^{a}_2 = W^p_2 (T^*)^T, \quad b^{a}_1 = T^* b^p_1.
\end{equation}
This holistic mathematical recovery rigorously restores the exact parameter arrangement of the MLP block.

\subsection{Theoretical Error Bound of AGA}
To rigorously validate the effectiveness of the AGA, we provide a theoretical upper bound for the recovery error in the continuous transformations of attention modules, alongside a strict exact-recovery guarantee for the discrete permutations in MLP blocks.

\begin{theorem}[Error Bound of Attention Module Recovery]
\label{thm:attention_recovery}
If the recovery matrix $T^*$ is obtained via the least-squares objective $\min_{T} \| W^p T - W_{pre} \|_F^2$, the Frobenius norm of the recovery error $\mathcal{E} = \| W^a - W_{ft} \|_F$ is strictly upper-bounded by the magnitude of the task vector $\tau$:
\begin{equation}
    \mathcal{E} \le \| \tau \|_F
\end{equation}
\end{theorem}

\begin{theorem}[Error Bound of MLP Layer Recovery]
\label{thm:mlp_recovery}
We define the permutation margin of the pretrained model as $\delta_{min} = \min_{S \in \mathcal{P}, S \neq I} \| S W_{pre} - W_{pre} \|_F$, where $\mathcal{P}$ denotes the set of all valid permutation matrices.
This margin represents the minimum distance between $W_{pre}$ and any distinct permuted state of itself.
If the magnitude of the task vector satisfies $\| \tau \|_F < \frac{1}{2} \delta_{min}$, the linear sum assignment problem is mathematically guaranteed to output the exact inverse permutation $T^* = P^T$, resulting in strictly zero recovery error: $W^a = W_{ft}$.
\end{theorem}

In summary, Theorems~\ref{thm:attention_recovery} and ~\ref{thm:mlp_recovery} collectively establish a rigorous mathematical foundation for the Anchor-Guided Attack (AGA). By exploiting the inherent magnitude disparity of task arithmetic, we theoretically guarantee that AGA achieves strictly bounded recovery error against continuous protections and zero-error exact recovery against discrete permutations. Due to space constraints, the complete and detailed mathematical proofs for both theorems are deferred to the Supplementary Material.

\section{Experiments}
\subsection{Experiment Setups}
\label{Attack Experiment Setups}
\textbf{Models and Datasets.} 
We conduct experiments on visual classification, natural language processing, and natural language generation tasks.
For visual classification, we use two variants of the CLIP~\cite{radford2021learning}—ViT-B/32 and ViT-L/14—and evaluate on eight datasets following prior work~\cite{lu2024twin, chen2025defending, junhao2025disrupting}: SUN397~\cite{xiao2010sun}, Cars~\cite{krause20133d}, RESISC45~\cite{cheng2017remote}, EuroSAT~\cite{helber2019eurosat}, SVHN~\cite{netzer2011reading}, GTSRB~\cite{stallkamp2011german}, MNIST~\cite{deng2012mnist}, and DTD~\cite{cimpoi2014describing}.
For natural language processing tasks, we adopt GPT-2~\cite{radford2019language} as the backbone, and evaluate on eight tasks from the GLUE benchmark~\cite{wang2018glue}: CoLA~\cite{warstadt2019neural}, SST-2~\cite{socher2013recursive}, MRPC~\cite{dolan2005automatically}, STS-B~\cite{cer2017semeval}, QQP~\cite{le2021improve}, MNLI~\cite{williams2017broad}, QNLI~\cite{rajpurkar2016squad}, and RTE~\cite{giampiccolo2007third}.
For natural language generation, we use Qwen2-7B~\cite{yang2024qwen2technicalreport} as the backbone and evaluate on AlpacaEval~\cite{li2023alpacaeval} for instruction following, GSM8K~\cite{cobbe2021training} for mathematical reasoning, and MBPP~\cite{austin2021program} for program synthesis.
Additional details are provided in the supplementary material.

\begin{table}[t]
    \centering
    \caption{
    Evaluation of protected-task performance under various defense mechanisms and our AGA framework using the ViT-B/32 backbone and various merging methods.
    {\color{Red} {$\uparrow \Delta$ }} presents the accuracy recovered by AGA relative to the only-protected setting. 
    } 
    \label{tab:ViT-B32}
    \setlength{\tabcolsep}{1.2pt} 
    \renewcommand{\arraystretch}{1.5} 
    
    \newcommand{\upR}[1]{\,\makebox[0pt][l]{\raisebox{-0ex}{\scalebox{1}{(\textcolor{red}{$\uparrow$ #1})}}}}
    \newcommand{\headerDelta}{\,\makebox[0pt][l]{\raisebox{-0ex}{\scalebox{1}{({\color{Red} {$\uparrow \Delta$ }})}}}}
    \resizebox{\textwidth}{!}{
        \begin{NiceTabular}{
            l 
            c 
            *{8}{>{\centering\arraybackslash}p{1.4cm}} 
            c@{\hskip 33pt} 
        }
            \hline
            Protect & Attack & SUN397 & Cars & RESISC45 & EuroSAT & SVHN & GTSRB & MNIST & DTD & Avg\headerDelta \\
            \hline
            
            % === Task Arithmetic ===
            \rowcolor{gray!10}\Block{1-11}{\textbf{Task Arithmetic~\cite{ilharco2022editing}}} & & & & & & & & & & \\
            \hline
            
             \Block[fill=white]{1-1}{-} & - & 61.75 & 55.52 & 60.24 & 69.32 & 83.25 & 63.20 & 97.55 & 47.02 & 67.23 \\
            \hline
            
            \Block[fill=white]{2-1}{Params} & - & 0.62  & 0.62  & 4.49  & 13.44 & 7.89  & 3.56  & 10.24 & 3.09  & 5.49  \\[-1.4ex]
             & AGA & 61.51 & 55.15 & 59.94 & 68.40 & 81.13 & 61.46 & 97.06 & 46.44 & 66.39\upR{60.90} \\
            \hline
            
            \Block[fill=white]{2-1}{Params-D} & - & 0.47  & 0.63  & 4.34  & 13.72 & 8.05  & 3.54  & 10.28 & 2.98  & 5.50  \\[-1.4ex]
             & AGA & 61.05 & 54.67 & 59.14 & 65.42 & 79.00 & 59.13 & 96.18 & 45.27 & 64.98\upR{59.48} \\
            \hline
            
            \Block[fill=white]{2-1}{MergeLock} & - & 0.30  & 0.50  & 2.99  & 12.42 & 9.05  & 3.48  & 10.28 & 1.76  & 5.10  \\[-1.4ex]
             & AGA & 61.50 & 55.17 & 59.94 & 68.40 & 81.15 & 61.45 & 97.06 & 46.44 & 66.39\upR{61.29} \\
            \hline
            
            \Block[fill=white]{2-1}{MergeBarrier} & - & 26.01 & 10.04 & 19.19 & 22.00 & 45.10 & 18.50 & 70.59 & 23.09 & 29.32 \\[-1.4ex]
             & AGA & 58.56 & 52.39 & 56.82 & 62.18 & 73.79 & 53.70 & 86.09 & 42.89 & 60.80\upR{31.48} \\
            \hline
             
            \rowcolor{gray!10} \Block{1-11}{\textbf{CAT Merging~\cite{sun2025cat}}} & & & & & & & & & & \\
            \hline
            
             \Block[fill=white]{1-1}{-} & - & 65.80 & 60.42 & 69.47 & 80.38 & 83.03 & 59.47 & 98.47 & 50.43 & 70.93 \\
            \hline
            
            \Block[fill=white]{2-1}{Params} & - & 0.81  & 0.42  & 3.54  & 14.70 & 8.97  & 2.57  & 10.30 & 2.02  & 5.42  \\[-1.42ex]
             & AGA & 65.25 & 59.79 & 68.03 & 79.80 & 81.74 & 58.13 & 98.35 & 49.52 & 70.08\upR{64.66} \\
            \hline
            
            \Block[fill=white]{2-1}{Params-D} & - & 0.72  & 0.41  & 3.60  & 14.81 & 9.00  & 2.88  & 10.19 & 2.37  & 5.50  \\[-1.42ex]
             & AGA & 65.33 & 59.20 & 67.09 & 78.37 & 80.22 & 56.85 & 97.51 & 47.36 & 68.99\upR{63.49} \\
            \hline
            
            \Block[fill=white]{2-1}{MergeLock} & - & 0.33  & 0.40  & 2.11  & 12.92 & 13.69 & 5.08  & 10.27 & 2.34  & 5.89  \\[-1.42ex]
             & AGA & 65.38 & 60.10 & 68.83 & 80.12 & 81.06 & 55.78 & 98.02 & 49.57 & 69.86\upR{63.97} \\
            \hline
            
            \Block[fill=white]{2-1}{MergeBarrier} & - & 22.83 & 5.78  & 22.90 & 31.56 & 50.70 & 12.88 & 80.25 & 26.81 & 31.71 \\[-1.42ex]
            & AGA & 63.90 & 27.19 & 66.00 & 76.25 & 75.48 & 51.04 & 85.28 & 43.10 & 61.03\upR{29.32} \\
            \hline
             
            \rowcolor{gray!10} \Block{1-11}{\textbf{LOT Merging~\cite{wanglocalizing}}} & & & & & & & & & & \\
            \hline
            
            \Block[fill=white]{1-1}{-} & - & 66.96 & 65.70 & 75.65 & 92.80 & 92.65 & 82.98 & 98.85 & 57.66 & 79.16 \\
            \hline
            
            \Block[fill=white]{2-1}{Params} & - & 0.51  & 0.50  & 3.18  & 9.50  & 11.22 & 4.70  & 10.28 & 4.04  & 5.49  \\[-1.4ex]
             & AGA & 66.98 & 65.27 & 73.26 & 92.38 & 91.43 & 82.88 & 98.42 & 57.18 & 78.48\upR{72.99} \\
            \hline
            
            \Block[fill=white]{2-1}{Params-D} & - & 0.57  & 0.56  & 3.01  & 10.16 & 11.09 & 4.41  & 10.36 & 3.90  & 5.51  \\[-1.4ex]
             & AGA & 66.29 & 64.74 & 73.01 & 90.50 & 90.16 & 81.35 & 97.84 & 56.58 & 77.56\upR{72.05} \\
            \hline
            
            \Block[fill=white]{2-1}{MergeLock} & - & 0.26  & 0.62  & 4.20  & 12.04 & 6.77  & 2.34  & 10.77 & 1.76  & 4.85  \\[-1.4ex]
             & AGA & 66.98 & 65.14 & 74.66 & 91.32 & 91.62 & 81.08 & 98.35 & 57.70 & 78.36\upR{73.51} \\
            \hline
            
            \Block[fill=white]{2-1}{MergeBarrier} & - & 17.74 & 4.32  & 13.00 & 12.42 & 32.86 & 15.60 & 10.20 & 17.07 & 15.40 \\[-1.4ex]
             & AGA & 63.48 & 60.20 & 69.90 & 85.09 & 82.61 & 66.22 & 87.30 & 49.80 & 70.58\upR{55.18} \\
            \hline
             
        \end{NiceTabular}%
    }
\end{table}

\textbf{Metrics.} 
For all image classification and text classification tasks, we report the Top-1 accuracy. 
For natural language generation tasks evaluated on Qwen2-7B, we adopt domain-specific metrics corresponding to their benchmarks: win rate for AlpacaEval, exact match accuracy for GSM8K, and pass@1 for MBPP.

\textbf{Merging Methods.}
To comprehensively demonstrate the generality of our AGA, we evaluate its effectiveness across a range of model merging paradigms, ranging from the classic baseline of Task Arithmetic (TA)~\cite{ilharco2022editing} to the latest SOTA advancements, specifically CAT Merging~\cite{sun2025cat} and LOT Merging~\cite{wanglocalizing}.

\textbf{Defense Baselines.} We evaluate the robustness of our AGA against four established parameter-space protection mechanisms: Params~\cite{junhao2025disrupting}, Params-D~\cite{junhao2025disrupting}, MergeLock~\cite{wang2025model}, and MergeBarrier~\cite{li2025not}. 
The formal mathematical definitions and operational mechanics of these defenses are previously detailed in Sec.~\ref{Existing Defense Mechanisms}. 

\textbf{Evaluation Strategy.} 
To simulate a realistic zero-knowledge adversary, our evaluation follows three core protocols:
Defense-Agnostic Setting: Since adversaries typically cannot identify which specific models are protected, we blindly apply our AGA method to all candidate models.
\textit{Protected-Task Performance:} This metric quantifies AGA's ability to restore the protected task-specific expertise.
Specifically, in a scenario where only one model is protected while others are unprotected, we report the performance of the final merged model on the dataset corresponding to that explicitly protected model.
\textit{Cross-Task Average Utility:} This metric measures the average performance of the merged model across all evaluated datasets. 

\begin{table}[t]
    \centering
    \caption{Evaluation of protected-task performance under various defense mechanisms and our AGA framework using the ViT-L/14 backbone. 
    {\color{Red} {$\uparrow \Delta$ }} presents the accuracy recovered by AGA relative to the only-protected setting.
    } 
    \label{tab:ViT-L14_supp}
    
    \setlength{\tabcolsep}{1.2pt} 
    \renewcommand{\arraystretch}{1.5} 
    
    \newcommand{\upR}[1]{\,\makebox[0pt][l]{\raisebox{-0ex}{\scalebox{1}{(\textcolor{red}{$\uparrow$ #1})}}}}
    \newcommand{\headerDelta}{\,\makebox[0pt][l]{\raisebox{-0ex}{\scalebox{1}{({\color{Red} {$\uparrow \Delta$ }})}}}}

    \resizebox{\textwidth}{!}{% 
        \begin{NiceTabular}{
            l 
            c 
            *{8}{>{\centering\arraybackslash}p{1.4cm}} 
            c@{\hskip 35pt}
        }
            \hline
            Protect & Attack & SUN397 & Cars & RESISC45 & EuroSAT & SVHN & GTSRB & MNIST & DTD & Avg\headerDelta \\
            \hline
            
            \Block[fill=white]{1-1}{-} & - & 73.83 & 82.09 & 87.60 & 93.76 & 87.91 & 86.78 & 98.94 & 65.69 & 84.58 \\
            \hline
            
            \Block[fill=white]{2-1}{Params~\cite{junhao2025disrupting}} & - & 0.39  & 0.46  & 4.14  & 21.36 & 7.96  & 2.14  & 9.82  & 2.13  & 6.05  \\[-1.4ex]
             & AGA & 73.36 & 82.02 & 85.96 & 93.82 & 83.14 & 80.36 & 98.27 & 64.84 & 82.72\upR{76.67} \\
            \hline
            
            \Block[fill=white]{2-1}{Params-P~\cite{junhao2025disrupting}} & - & 0.31  & 0.52  & 4.00  & 19.45 & 7.58  & 2.29  & 9.88  & 2.02  & 5.76  \\[-1.4ex]
             & AGA & 72.44 & 81.11 & 84.77 & 92.10 & 82.86 & 80.23 & 98.06 & 64.33 & 81.99\upR{76.23} \\
            \hline
            
            \Block[fill=white]{2-1}{MergeLock~\cite{wang2025model}} & - & 0.32  & 0.66  & 1.86  & 10.20 & 9.52  & 2.11  & 7.66  & 1.81  & 4.27  \\[-1.4ex]
             & AGA & 73.36 & 82.02 & 85.96 & 93.82 & 83.14 & 80.35 & 98.27 & 64.84 & 82.72\upR{78.45} \\
            \hline
            
            \Block[fill=white]{2-1}{MergeBarrier~\cite{li2025not}} & - & 5.59  & 0.83  & 9.00  & 9.76  & 7.65  & 4.20  & 6.60  & 12.39 & 7.00  \\[-1.4ex]
             & AGA & 71.16 & 78.32 & 83.60 & 87.42 & 84.01 & 78.20 & 95.32 & 62.36 & 80.05\upR{73.05} \\
            \hline
             
        \end{NiceTabular}%
    }
\end{table}

\begin{table}[t]
    \centering
    \caption{Evaluation of protected-task performance under various defense mechanisms and our AGA framework using the Qwen2-7B backbone.
    {\color{Red} {$\uparrow \Delta$ }} presents the accuracy recovered by AGA relative to the only-protected setting. 
    } 
    \label{tab:Qwen}
    \setlength{\tabcolsep}{1.2pt} 
    \renewcommand{\arraystretch}{1.4} 

    \newcommand{\upR}[1]{\,\makebox[0pt][l]{\raisebox{-0ex}{\scalebox{1}{(\textcolor{red}{$\uparrow$ #1})}}}}
    \newcommand{\headerDelta}{\,\makebox[0pt][l]{\raisebox{-0ex}{\scalebox{1}{({\color{Red} {$\uparrow \Delta$ }})}}}}

    \resizebox{0.55\textwidth}{!}{% 
        \begin{NiceTabular}{
            l 
            c 
            *{3}{>{\centering\arraybackslash}p{1.3cm}} 
            c@{\hskip 30pt} 
        }
            \hline
            Protect & Attack & Alpaca & GSM8K & MBPP & Avg\headerDelta \\
            \hline
            
            \Block[fill=white]{1-1}{-} & - & 29.04 & 66.79 & 51.48 & 49.10 \\
            \hline
            
            \Block[fill=white]{2-1}{Params~\cite{junhao2025disrupting}} & - & 6.34  & 1.74  & 1.17  & 3.08  \\[-1.4ex]
             & AGA & 29.01 & 66.38 & 51.40 & 48.93\upR{45.85} \\
            \hline
            
            \Block[fill=white]{2-1}{Params-D~\cite{junhao2025disrupting}} & - & 6.12  & 1.79  & 1.29  & 3.07  \\[-1.4ex]
             & AGA & 28.59 & 64.11 & 50.68 & 47.79\upR{44.72} \\
            \hline
            
            \Block[fill=white]{2-1}{MergeLock~\cite{wang2025model}} & - & 3.25  & 2.12  & 1.42  & 2.26  \\[-1.4ex]
             & AGA & 29.03 & 66.54 & 51.46 & 49.01\upR{46.75} \\
            \hline
            
            \Block[fill=white]{2-1}{MergeBarrier~\cite{li2025not}} & - & 14.51 & 19.71 & 13.17 & 15.80 \\[-1.4ex]
             & AGA & 27.49 & 63.77 & 50.02 & 47.09\upR{31.29} \\
            \hline
             
        \end{NiceTabular}%
    }
\end{table}

\subsection{Experimental Results}
In this section, we systematically evaluate the efficacy of our proposed Anchor-Guided Attack (AGA) against SOTA defenses. 
Unless stated otherwise, all merging operations utilize Task Arithmetic~\cite{ilharco2022editing} as the merging paradigm.
As evidenced in Table~\ref{tab:ViT-B32} and Table~\ref{tab:ViT-L14_supp}, AGA consistently defeats SOTA protections, recovering the accuracy to near-unprotected levels across all merging strategies.
For instance, under LOT Merging~\cite{wanglocalizing}, MergeLock suppresses average protected-task performance from 79.16\% to a mere 4.85\%. 
Remarkably, AGA successfully restores performance to 78.36\%, a negligible gap of less than 1\% from the unprotected baseline. 
Such near-complete recoveries across diverse merging paradigms confirm AGA's robust adaptability to different parameter-fusion algorithms.
Beyond vision models, AGA demonstrates formidable capabilities across NLP (GPT-2 in Table~\ref{tab:GPT2_supp} in the \textbf{supplementary material}) and large-scale text generation (Qwen2-7B in Table~\ref{tab:Qwen}) benchmarks. 
On complex Qwen2-7B generation tasks, Params-D drastically degrades the average performance to 3.07\%, whereas AGA successfully reconstructs the weights to achieve a score of 47.79\%.

\begin{table}[t]
    \centering
    \caption{Cross-task average utility of merged ViT-B/32 models under baseline defenses and our AGA attack.
    {\color{Red} {$\uparrow \Delta$ }} presents the accuracy recovered by AGA relative to the only-protected setting. 
    } 
    \label{tab:avg}
    \setlength{\tabcolsep}{1.2pt} 
    \renewcommand{\arraystretch}{1.5} 

    \newcommand{\upR}[1]{\,\makebox[0pt][l]{\raisebox{-0ex}{\scalebox{1}{(\textcolor{red}{$\uparrow$ #1})}}}}
    \newcommand{\headerDelta}{\,\makebox[0pt][l]{\raisebox{-0ex}{\scalebox{1}{({\color{Red} {$\uparrow \Delta$ }})}}}}

    \resizebox{\textwidth}{!}{%
        \begin{NiceTabular}{
            l 
            c 
            *{8}{>{\centering\arraybackslash}p{1.4cm}} 
            c@{\hskip 35pt}
        }
            \hline
            Protect & Attack & SUN397 & Cars & RESISC45 & EuroSAT & SVHN & GTSRB & MNIST & DTD & Avg\headerDelta \\
            \hline
            
            \Block[fill=white]{1-1}{-} & - & 67.23 & 67.23 & 67.23 & 67.23 & 67.23 & 67.23 & 67.23 & 67.23 & 67.23 \\
            \hline
            
            \Block[fill=white]{2-1}{Params~\cite{junhao2025disrupting}} & - & 5.74  & 5.92  & 5.50  & 5.71  & 5.05  & 6.21  & 5.84  & 5.98  & 5.74  \\[-1.4ex]
             & AGA & 67.19 & 67.17 & 67.19 & 67.12 & 67.12 & 67.02 & 67.38 & 67.17 & 67.17\upR{61.43} \\
            \hline
            
            \Block[fill=white]{2-1}{Params-D~\cite{junhao2025disrupting}} & - & 5.68  & 6.09  & 5.49  & 5.73  & 5.46  & 5.78  & 6.67  & 6.09  & 5.87 \\[-1.4ex]
             & AGA & 67.25 & 67.19 & 67.08 & 66.82 & 67.36 & 66.89 & 67.88 & 67.04 & 67.19\upR{61.32} \\
            \hline
            
            \Block[fill=white]{2-1}{MergeLock~\cite{wang2025model}} & - & 3.84  & 5.02  & 4.09  & 5.27  & 4.99  & 5.75  & 4.60  & 5.79  & 4.92 \\[-1.4ex]
             & AGA & 67.19 & 67.18 & 67.19 & 67.12 & 67.11 & 67.02 & 67.38 & 67.17 & 67.17\upR{62.25} \\
            \hline
            
            \Block[fill=white]{2-1}{MergeBarrier~\cite{li2025not}} & - & 38.68 & 38.74 & 38.31 & 37.20 & 34.60 & 37.17 & 36.20 & 38.70 & 37.45 \\[-1.4ex]
             & AGA & 67.17 & 67.19 & 66.15 & 64.85 & 67.34 & 66.15 & 69.30 & 66.47 & 66.83\upR{29.38} \\
            \hline
             
        \end{NiceTabular}%
    }
\end{table}

\begin{table}[t]
    \centering
    \caption{
    Evaluation of protected-task performance under combined defense and AGA using the ViT-B/32 backbone. 
    Notably, \textbf{P} represents Params~\cite{junhao2025disrupting}, \textbf{L} represents MergeLock~\cite{wang2025model} and \textbf{B} represents MergeBarrier~\cite{li2025not}.
    {\color{Red} {$\uparrow \Delta$ }} presents the accuracy recovered by AGA relative to the only-protected setting. } 
    \label{tab:pairwise}
    
    \setlength{\tabcolsep}{1.2pt} 
    \renewcommand{\arraystretch}{1.5} 
    
    \newcommand{\upR}[1]{\,\makebox[0pt][l]{\raisebox{-0ex}{\scalebox{1}{(\textcolor{red}{$\uparrow$ #1})}}}}
    \newcommand{\headerDelta}{\,\makebox[0pt][l]{\raisebox{-0ex}{\scalebox{1}{({\color{Red} {$\uparrow \Delta$ }})}}}}

    \resizebox{\textwidth}{!}{%
        \begin{NiceTabular}{
            l 
            c 
            *{8}{>{\centering\arraybackslash}p{1.4cm}} 
            c@{\hskip 35pt}
        }
            \hline
            Protect & Attack & SUN397 & Cars & RESISC45 & EuroSAT & SVHN & GTSRB & MNIST & DTD & Avg\headerDelta \\
            \hline
            
            \Block[fill=white]{1-1}{-} & - & 61.75 & 55.52 & 60.24 & 69.32 & 83.25 & 63.20 & 97.55 & 47.02 & 67.23 \\
            \hline
            
            \Block[fill=white]{2-1}{\textbf{P} \& \textbf{L}} & - & 0.26  & 0.57  & 3.96  & 9.10  & 9.68  & 1.77  & 9.83  & 2.66  & 4.73  \\[-1.4ex]
             & AGA & 61.50 & 55.15 & 59.93 & 68.42 & 81.15 & 61.45 & 97.06 & 46.44 & 66.39\upR{61.66} \\
            \hline
            
            \Block[fill=white]{2-1}{\textbf{P} \& \textbf{B}} & - & 11.16 & 1.63  & 11.25 & 20.62 & 9.71  & 3.99  & 15.62 & 16.70 & 11.34 \\[-1.4ex]
             & AGA & 58.56 & 52.39 & 56.82 & 62.18 & 73.79 & 52.70 & 86.09 & 42.89 & 60.68\upR{49.34} \\
            \hline
            
            \Block[fill=white]{2-1}{\textbf{L} \& \textbf{B}} & - & 0.30  & 0.60  & 2.20  & 9.86  & 9.25  & 2.72  & 7.48  & 2.02  & 4.30 \\[-1.4ex]
             & AGA & 58.56 & 52.37 & 56.82 & 62.16 & 73.70 & 52.72 & 86.13 & 42.84 & 60.66\upR{56.36} \\
            \hline

            \Block[fill=white]{2-1}{\textbf{P} \& \textbf{L} \& \textbf{B}} & - & 0.27  & 0.61  & 2.09  & 9.44  & 9.10  & 2.58  & 7.25  & 1.94  & 4.16 \\[-1.4ex]
             & AGA & 58.54 & 52.39 & 56.81 & 62.16 & 73.70 & 52.77 & 86.11 & 42.82 & 60.66\upR{56.50} \\
            \hline
        \end{NiceTabular}%
    }
\end{table}

Moreover, AGA bypasses targeted protections without compromising the global integrity of the multi-task model. 
As demonstrated in Table~\ref{tab:avg}, while MergeLock suppresses the cross-task average utility to 4.92\%, AGA restores it to 67.17\%, missing the original unprotected performance (67.23\%) by a mere 0.06\%. 
This empirical evidence validates that AGA achieves high-fidelity parameter reconstruction, ensuring that recovered models maintain full functional compatibility with the original task-vector space.
Furthermore, \textbf{supplementary} Table~\ref{tab:standalone_attack_supp} confirms that AGA achieves high-fidelity recovery regardless of the initial protection status.
The standalone performance of both protected and unprotected models post-attack exhibits a strictly negligible degradation compared to their original baselines.

MergeBarrier alters MLP topology via Taylor expansion, rendering standard inversion mathematically ill-posed. 
To circumvent this, AGA bypasses the modified blocks and reverts to pretrained MLP layers during merging. 
Despite MergeBarrier's naturally weaker baseline (e.g., 26.01\% on SUN397), AGA significantly elevates the accuracy to 58.56\%, as shown in Table~\ref{tab:ViT-B32}. 
While missing the task-specific MLP fine-tuning yields slightly lower performance than purely parameter-space defenses, successfully inverting the attention modules proves AGA's critical robustness against structural alterations.

To simulate a worst-case scenario, we evaluate AGA against composite defenses that apply multiple protections simultaneously. 
As Table~\ref{tab:pairwise} illustrates, when combining purely parameter-level protections (e.g., P and L), AGA recovers the average accuracy to 66.39\%. 
This closely matches the 67.23\% unprotected baseline, leaving a marginal gap under 1\%. 
In contrast, integrating MergeBarrier (e.g., P \& B or P \& L \& B) noticeably decreases the recovered accuracy to approximately 60.66\%. 
This reduction occurs because MergeBarrier fundamentally alters MLP architectures, inherently limiting perfect parameter-space reconstruction.
Ultimately, these results confirm AGA's effectiveness, proving that simply concatenating existing methods fails to provide robust security.

\section{Countermeasure}
\subsection{Anchor-Repulsive Fine-tuning Protection}
Our theoretical analysis in Sec.~\ref{Attack Method} reveals that the effectiveness of AGA hinges on a critical assumption: the magnitude of the protected task vector is disproportionately small compared to the pretrained anchor. 
Therefore, to fundamentally defend against this inversion attack, the defender must systematically violate this assumption by artificially expanding the task vector's magnitude. 
Because the task vector is inherently formed during the finetuning phase, this theoretical insight naturally motivates us to implement protective interventions directly during the fine-tuning process.

The recent proactive defense MergeGuard~\cite{chen2025defending} also adopts tuning-stage protection, attempting to disperse task weights via global $L_2$ regularization to intentionally aggravate parameter conflicts and induce destructive interference between different tasks.
However, this approach exhibits two critical flaws. 
First, an overall-model constraint inherently restricts the parameter space, inevitably degrading the protected model's standalone capabilities. 
Second, our empirical evaluations reveal that MergeGuard~\cite{chen2025defending} remains vulnerable to AGA across several datasets, as shown in Table~\ref{tab:TA-B32}. 
Therefore, merely dispersing weights fails to mathematically guarantee a sufficient magnitude expansion.

These dual vulnerabilities highlight a crucial design imperative: an effective defense must decisively expand the target parameter distance without polluting the broader network. 
Building upon this insight, we introduce Anchor-Repulsive Fine-tuning (ARF). 
Instead of penalizing the entire network, ARF surgically applies an adaptive Euclidean repulsive force exclusively to the attention projection matrices.
This localized intervention is deliberate: their continuous inversion relies on minimizing the Euclidean distance (bounded by $\|\tau\|_F$, as shown in Theorem~\ref{thm:attention_recovery}), making them highly susceptible to distance-based repulsion.
Conversely, the discrete extraction of MLP layers relies on scale-invariant cosine matching, rendering Euclidean expansion ineffective. 

Therefore, during the fine-tuning phase, we introduce a distance-based margin penalty that actively pushes only these designated attention parameters away from their pretrained anchors until they reach a predefined safety boundary:
\begin{equation}
\label{loss}
    L_{total} = L_{CE} + \lambda_{dist} \sum_{\theta \in \{W_Q, W_K, W_V, W_O\}} \min(                           0, \rho \| \theta_{pre} \|_2- \| \theta - \theta_{pre} \|_2)
\end{equation}
where $L_{CE}$ is the standard cross-entropy task loss, $\theta$ represents the targeted attention weight, $\theta_{pre}$ is the corresponding pretrained anchor, $\lambda_{dist}$ controls the strength of the repulsive force.
$\rho$ denotes a predefined expansion ratio that establishes the relative safety margin based on the anchor's original magnitude.
By employing this margin-aware repulsion, ARF actively enlarges the distance $\| \theta - \theta_{pre} \|_2$ only when necessary, drastically expanding the magnitude of the targeted task vectors without polluting the representation capacity of other layers. 
Once the model is fine-tuned to satisfy this margin, we apply the standard invertible matrix transformations to protect the attention modules.

\begin{figure}[t]
\captionsetup[subfigure]{font=small}
  \centering
  \setlength{\abovecaptionskip}{0.2cm} 
  
  \begin{subfigure}[c]{0.48\textwidth}
    \centering
    \includegraphics[width=\linewidth]{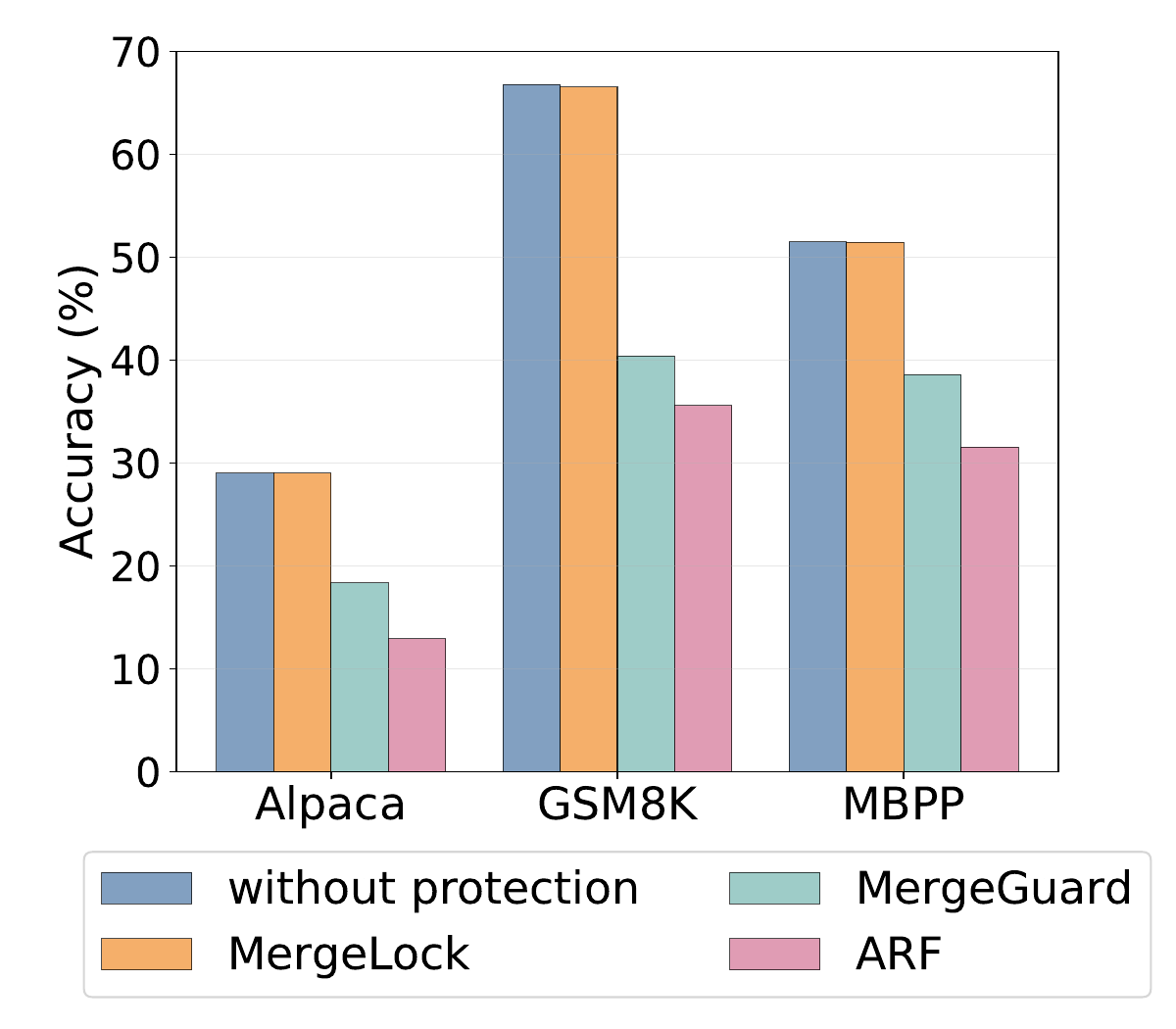}
    \caption{Protection efficacy under attack.}
    \label{fig: Qwen_protect(a)}
  \end{subfigure}
  \begin{subfigure}[c]{0.48\textwidth}
    \centering
    \includegraphics[width=\linewidth]{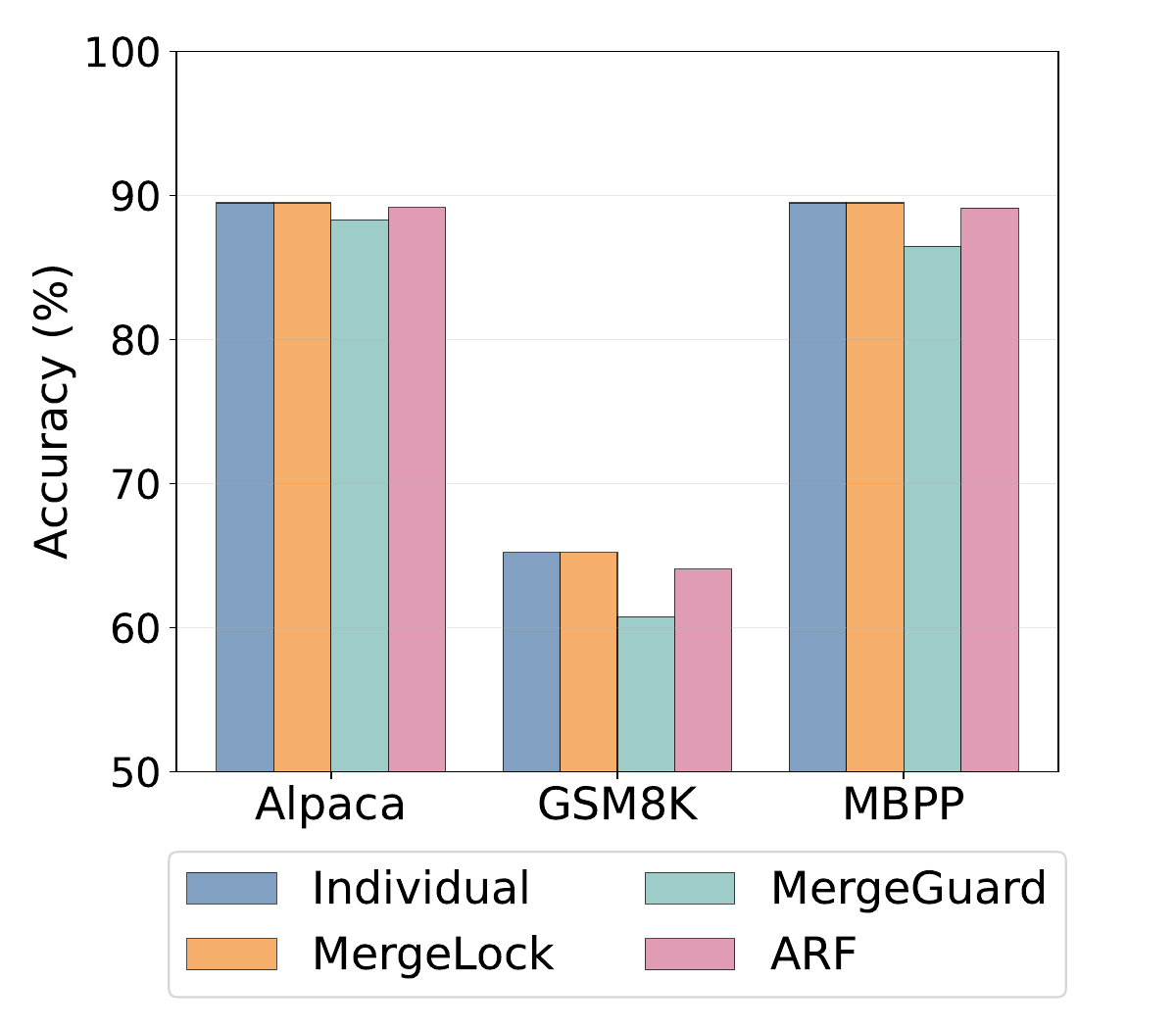}
    \caption{Standalone utility of protection.}
    \label{fig: Qwen_protect(b)}
  \end{subfigure}

  \caption{Evaluation of protection efficacy against attack and standalone utility on Qwen2-7B. 
  (a) Protected-task performance of AGA-attacked models under various defenses. 
  (b) Standalone accuracy of individual fine-tuned models.}
  \label{fig:combined_analysis}
\end{figure}

\subsection{Experiment}
\textbf{Experiment Setups}.
The experimental configurations strictly follow the protocols in Sec.~\ref{Attack Experiment Setups}. 
Across all datasets, the defense hyperparameters in Eq.~\ref{loss} are set to $\lambda_{dist} = 1.0$ and $\rho = 0.05$. 
Because ARF proactively expands parameter magnitude during fine-tuning, it is inherently orthogonal to post-training linear protections.
Consequently, "ARF" in our results denotes the complete defensive pipeline: applying our repulsive fine-tuning followed by standard invertible matrix protection on attention modules. 
We compare ARF against MergeGuard~\cite{chen2025defending}, a state-of-the-art tuning-stage defense, focusing on two dimensions: \textit{Protection Capability} (resisting AGA and degrading unauthorized merged models) and \textit{Standalone Utility} (preserving original task accuracy).

\textbf{Experiment Results.} 
As demonstrated in Fig.~\ref{fig: Qwen_protect(a)} and Table~\ref{tab:TA-B32}, ARF exhibits superior defensive capabilities across modalities. 
Without attacks, ARF consistently suppresses unauthorized merging utility to near-random levels. 
For instance, on the ViT-B/32 backbone in Table~\ref{tab:TA-B32}, ARF reduces the merged accuracy to a mere 0.28\%, significantly outperforming MergeGuard's 43.52\%. 
Furthermore, ARF maintains its strong protective barrier even against our potent AGA method. 
Remarkably, the accuracy of ARF-protected models under attack remains substantially lower than that of MergeGuard in its unattacked state (see Table~\ref{tab:TA-B32}, Table~\ref{tab:TA-L14_supp}, and \textbf{supplementary} Table~\ref{tab:GPT-2_supp}). 
This validates that ARF's optimization objective effectively eliminates the exploitable parameter-space proximity. 
More results for CAT merging~\cite{sun2025cat} and LOT merging~\cite{wanglocalizing} on ViT-B/32 are detailed in Table~\ref{tab:CAT_protect_supp} and Table~\ref{tab:LOT_protect_supp} in the supplementary material.

\begin{table}[t]
    \centering
    \caption{
      Evaluation of protected-task performance under protection and AGA using the ViT-B/32 backbone.
    }
    \label{tab:TA-B32}
    
    \setlength{\tabcolsep}{1.2pt} 
    \renewcommand{\arraystretch}{1.5} 

    \resizebox{\textwidth}{!}{%
        \begin{NiceTabular}{
            l 
            c 
            *{8}{>{\centering\arraybackslash}p{1.4cm}} 
            c
        }
            \hline
            Protect & Attack & SUN397 & Cars & RESISC45 & EuroSAT & SVHN & GTSRB & MNIST & DTD & Avg \\
            \hline
            
            \Block[fill=white]{1-1}{-} & - & 61.75 & 55.52 & 60.24 & 69.32 & 83.25 & 63.20 & 97.55 & 47.02 & 67.23 \\
            \hline

            MergeLock~\cite{wang2025model} & \Block[fill=white]{3-1}{-} & 0.30  & 0.50  & 2.99  & 12.42 & 9.05  & 3.48  & 10.28 & 1.76  & 5.10 \\[-1.4ex]
            MergeGuard~\cite{chen2025defending} &  & 43.52 & 28.49 & 29.54 & 39.66 & 40.40  & 19.81 & 35.36 & 34.97 & 33.97 \\[-1.4ex]
            ARF &  & 0.28  & 0.52  & 2.18  & 10.09 & 8.20  & 3.07  & 10.44 & 2.29  & \textbf{4.63} \\
            \hline

            MergeLock~\cite{wang2025model} & \Block[fill=white]{3-1}{AGA} & 61.50 & 55.17 & 59.94 & 68.40 & 81.15 & 61.45 & 97.06 & 46.44 & 66.39 \\[-1.4ex]
            MergeGuard~\cite{chen2025defending} &  & 46.23 & 34.26 & 59.91 & 68.50 & 57.12 & 42.81 & 49.04 & 46.62 & 50.56 \\[-1.4ex]
            ARF &  & 38.18 & 23.05 & 17.75 & 34.67 & 37.73 & 14.38 & 29.12 & 24.49 & \textbf{27.42} \\
            \hline
        \end{NiceTabular}%
    }
\end{table}

\begin{table}[t]
    \centering
    \caption{Evaluation of protected-task performance under protection and AGA using the ViT-L/14 backbone.}
    \label{tab:TA-L14_supp}
    
    \setlength{\tabcolsep}{1.2pt} 
    \renewcommand{\arraystretch}{1.5}

    \resizebox{\textwidth}{!}{%
        \begin{NiceTabular}{
            l 
            c 
            *{8}{>{\centering\arraybackslash}p{1.4cm}} 
            c
        }
            \hline
            Protect & Attack & SUN397 & Cars & RESISC45 & EuroSAT & SVHN & GTSRB & MNIST & DTD & Avg \\
            \hline
            
            \Block[fill=white]{1-1}{-} & - & 73.83 & 82.09 & 87.60 & 93.76 & 87.91 & 86.78 & 98.94 & 65.69 & 84.58 \\
            \hline
            \Block[fill=white]{1-1}
            {MergeLock~\cite{wang2025model}} & \Block[fill=white]{3-1}{-} & 0.32  & 0.66  & 1.86  & 10.20 & 9.52  & 2.11  & 7.66  & 1.81  & 4.27 \\[-1.4ex]
            \Block[fill=white]{1-1}{MergeGuard~\cite{chen2025defending}} &  & 54.01 & 38.92 & 56.53 & 54.63 & 48.08 & 13.35 & 17.08 & 64.70 & 43.41 \\[-1.4ex]
            \Block[fill=white]{1-1}{ARF} &  & 0.31  & 0.85  & 1.94  & 11.07 & 9.98  & 2.40  & 7.43  & 2.25  & \textbf{4.53} \\
            \hline
            \Block[fill=white]{1-1}{MergeLock~\cite{wang2025model}} & \Block[fill=white]{3-1}{AGA} & 73.36 & 82.02 & 85.96 & 93.82 & 83.14 & 80.35 & 98.27 & 64.84 & 82.72 \\[-1.4ex]
            \Block[fill=white]{1-1}{MergeGuard~\cite{chen2025defending}} &  &  60.21 & 46.50 & 84.19 & 90.47 & 62.83 & 50.41 & 48.73 & 65.43 & 63.60 \\[-1.4ex]
             \Block[fill=white]{1-1}{ARF} &  & 39.21 & 31.78 & 33.40 & 34.19 & 37.61 & 8.07  & 13.65 & 36.19 & \textbf{29.26} \\
            \hline
        \end{NiceTabular}
    }
\end{table}

\begin{table}[t]
    \centering
    \caption{
    Standalone accuracy of individual fine-tuned models with and without protection methods using the ViT-B/32 backbone.
        }
    \label{tab:B32-finetune}
    
    \resizebox{\textwidth}{!}{%
        \begin{tabular}{lccccccccc}
            \toprule
            Method & ~SUN397~ & ~~Cars~~ & RESISC45 & ~EuroSAT & ~~SVHN~~ & ~~GTSRB~ & ~~MNIST~ & ~~~DTD~~ & ~~~Avg~~ \\
            \midrule
            % Pretrain    & 62.93 & 59.73 & 60.35 & 44.86 & 31.61 & 32.56 & 48.25 & 43.99 & 48.04 \\
            Individual  & 74.54 & 76.40 & 91.67 & 97.89 & 97.39 & 98.94 & 99.65 & 73.99 & 88.81 \\
            % \midrule
            MergeGuard~\cite{chen2025defending} & 63.82 & 71.89 & 90.97 & 97.96 & 96.02 & 96.17 & 99.41 & 74.26 & 86.31 \\
            ARF                & 72.81 & 75.58 & 90.35 & 98.16 & 97.27 & 98.13 & 99.06 & 73.54 & 88.11 \\
            \bottomrule
        \end{tabular}
    }
\end{table}

Unlike MergeGuard~\cite{chen2025defending}, which frequently suffers performance drops due to global constraints, ARF preserves task-specific expertise by applying a localized repulsive force exclusively to attention modules. 
As shown in Table~\ref{tab:B32-finetune}, ARF achieves results nearly identical to standard fine-tuning on ViT-B/32. 
Specifically, its average accuracy of 88.11\% closely matches the 88.81\% baseline, noticeably outperforming MergeGuard's 86.31\%.
Similar high-fidelity results are observed for Qwen2-7B in Fig.~\ref{fig: Qwen_protect(b)}, as well as ViT-L/14 and GPT-2 in Tables~\ref{tab:L14_finetune_supp} and \ref{tab:GPT2_finetune_supp}  of the \textbf{supplementary material}. 
Ultimately, by breaking the magnitude disparity exploited by AGA, ARF effectively prevents unauthorized merging while incurring only a negligible drop in standalone performance.

\section{Conclusion}
In this work, we systematically analyze current model merging defenses and reveal that protected fine-tuned models are overwhelmingly dominated by their public pretrained anchors due to the inherently small magnitude of task vectors.
Capitalizing on this observation, we propose Anchor-Guided Attack (AGA), a universal framework that aligns the protected model with the pretrained anchor to bypass previous defenses.
Specifically, AGA deploys least squares regression and the Hungarian algorithm to neutralize these protections in attention and MLP modules.
Extensive cross-modal evaluations demonstrate that AGA decisively bypasses individual and composite defenses. 
We subsequently propose Anchor-Repulsive Fine-tuning (ARF) as an adaptive countermeasure to mitigate this specific threat.
The complete success of AGA demonstrates that linear weight transforms offer merely an illusion of security, urging the research community to develop fundamentally robust and training-aware paradigms.

\section*{Acknowledgments}
This work was funded by the National Natural Science Foundation of China under Grants (62276256, U2441251), the Beijing Natural Science Foundation (Z260008), and the National Key Research and Development Program of China (2026ZD1500301).

\bibliographystyle{splncs04}
\bibliography{main}

\clearpage
\setcounter{page}{1}
\section{The Proof}
\label{The theoretical proof}

\begin{theorem}[Error Bound of Attention Module Recovery]
\label{thm:attention_recovery_supp}
Let $W_{ft} = W_{pre} + \tau \in \mathbb{R}^{N \times D}$ (where $N \ge D$ and has full column rank) be the fine-tuned weight matrix, and $W^p = W_{ft} P$ be the protected model under a secret invertible transformation $P \in \mathbb{R}^{D \times D}$. 
If the recovery matrix $T^*$ is obtained via the least-squares objective $\min_{T} \| W^p T - W_{pre} \|_F^2$, the Frobenius norm of the recovery error $\mathcal{E} = \| W^a - W_{ft} \|_F$ is strictly upper-bounded by the magnitude of the task vector $\tau$:
\begin{equation}
    \mathcal{E} \le \| \tau \|_F.
\end{equation}
\end{theorem}

Proof. 

Given the closed-form solution of the least-squares objective:
\begin{equation}
    T^* = ((W^p)^T W^p)^{-1} (W^p)^T W_{pre}.
\end{equation}
We substitute the protection mechanism $W^p = W_{ft} P$ into the equation:
\begin{equation}
    T^* = (P^T W_{ft}^T W_{ft} P)^{-1} P^T W_{ft}^T W_{pre}.
\end{equation}
Using the property of invertible matrices, we expand the inverse term:
\begin{equation}
    T^* = P^{-1} (W_{ft}^T W_{ft})^{-1} (P^T)^{-1} P^T W_{ft}^T W_{pre} = P^{-1} (W_{ft}^T W_{ft})^{-1} W_{ft}^T W_{pre}.
\end{equation}
The recovered model $W^a$ is obtained by applying $T^*$ to the protected model $W^p$:

\begin{equation}
\begin{aligned}
        W^a = W^p T^* & = W_{ft} P \left[ P^{-1} (W_{ft}^T W_{ft})^{-1} W_{ft}^T W_{pre} \right]\\
        & = W_{ft} (W_{ft}^T W_{ft})^{-1} W_{ft}^T W_{pre}.
\end{aligned}
\end{equation}

Let $\Pi_{ft} = W_{ft} (W_{ft}^T W_{ft})^{-1} W_{ft}^T$. 
Mathematically, $\Pi_{ft}$ is the exact orthogonal projection matrix onto the column space of $W_{ft}$. 
Therefore, we have $W^a = \Pi_{ft} W_{pre}$.
Substituting the task arithmetic definition $W_{pre} = W_{ft} - \tau$:
\begin{equation}
    W^a = \Pi_{ft} (W_{ft} - \tau) = \Pi_{ft} W_{ft} - \Pi_{ft} \tau.
\end{equation}
Since $W_{ft}$ inherently lies within its own column space, projecting it onto itself leaves it unchanged, i.e., $\Pi_{ft} W_{ft} = W_{ft}$. 
Thus, the recovered model simplifies to:
\begin{equation}
    W^a = W_{ft} - \Pi_{ft} \tau.
\end{equation}
The recovery error matrix is exactly the projection of the task vector: $W^a - W_{ft} = - \Pi_{ft} \tau$.
Taking the Frobenius norm on both sides and applying the sub-multiplicative property of matrix norms:
\begin{equation}
    \mathcal{E} = \| -\Pi_{ft} \tau \|_F \le \| \Pi_{ft} \|_2 \| \tau \|_F.
\end{equation}
Because $\Pi_{ft}$ is an orthogonal projection matrix, its spectral norm (induced 2-norm) is strictly $\| \Pi_{ft} \|_2 = 1$. 
Consequently, the absolute recovery error satisfies:
\begin{equation}
    \mathcal{E} \le \| \tau \|_F.
\end{equation}

$\blacksquare$
Remark: This theorem mathematically proves why AGA is devastatingly effective.
This theorem mathematically proves why AGA is devastatingly effective. Crucially, because orthogonal matrices and non-zero diagonal matrices are fundamental subclasses of invertible matrices, this theoretical guarantee universally applies to defenses utilizing these specific structures (such as the diagonal transformations in Params or the orthogonal matrices in MergeLock).
The relative recovery error is bounded by $\frac{\| \tau \|_F}{\| W_{ft} \|_F} \approx \frac{\| \tau \|_F}{\| W_{pre} \|_F}$. 
As empirically observed, since the task vector magnitude is typically two to three orders of magnitude smaller than the pretrained anchor ($10^{-2}$ to $10^{-3}$), the theoretical maximum error of AGA is strictly bounded within this negligible margin. 
The defense is thus fundamentally dismantled by the geometry of the parameter space.

\begin{theorem}[Error Bound of MLP Layer Recovery]
\label{thm:mlp_recovery_supp}
We define the permutation margin of the pretrained model as $\delta_{min} = \min_{S \in \mathcal{P}, S \neq I} \| S W_{pre} - W_{pre} \|_F$, where $\mathcal{P}$ denotes the set of all valid permutation matrices.
This margin represents the minimum distance between $W_{pre}$ and any distinct permuted state of itself.
If the magnitude of the task vector satisfies $\| \tau \|_F < \frac{1}{2} \delta_{min}$, the linear sum assignment problem is mathematically guaranteed to output the exact inverse permutation $T^* = P^T$, resulting in strictly zero recovery error: $W^a = W_{ft}$.
\end{theorem}

Proof. The Hungarian Algorithm solving the LSAP aims to find a permutation matrix $T \in \mathcal{P}$ that optimally realigns $W^p$ with $W_{pre}$. 
Under standard Euclidean distance (which aligns with the optimization objective of cosine similarity for normalized vectors), this is equivalent to minimizing the global Frobenius distance:
\begin{equation}
    T^* = \arg\min_{T \in \mathcal{P}} \| T W^p - W_{pre} \|_F^2.
\end{equation}
Substitute the protection mechanism $W^p = P_{true} (W_{pre} + \tau)$ into the objective:
\begin{equation}
    \mathcal{J}(T) = \| T P_{true} (W_{pre} + \tau) - W_{pre} \|_F.
\end{equation}
Let $S = T P_{true} \in \mathcal{P}$. 
The optimization is equivalent to finding $S^*$ such that:
\begin{equation}
    S^* = \arg\min_{S \in \mathcal{P}} \| S (W_{pre} + \tau) - W_{pre} \|_F = \arg\min_{S \in \mathcal{P}} \| S W_{pre} - W_{pre} + S \tau \|_F.
\end{equation}
We evaluate the cost function $\mathcal{J}(S)$ under two conditions:
For the true inverse permutation ($S = I$, i.e., $T = P_{true}^T$):
\begin{equation}
    \mathcal{J}(I) = \| I W_{pre} - W_{pre} + I \tau \|_F = \| \tau \|_F.
\end{equation}
For any incorrect permutation ($S \neq I$):
Applying the reverse triangle inequality, we obtain:
\begin{equation}
    \mathcal{J}(S) = \| (S W_{pre} - W_{pre}) + S \tau \|_F \ge \| S W_{pre} - W_{pre} \|_F - \| S \tau \|_F.
\end{equation}
Since any permutation matrix is orthogonal, it perfectly preserves the Frobenius norm, meaning $\| S \tau \|_F = \| \tau \|_F$. 
Thus:
\begin{equation}
    \mathcal{J}(S) \ge \| S W_{pre} - W_{pre} \|_F - \| \tau \|_F.
\end{equation}
To guarantee that the exact inverse $S = I$ is the unique global minimum, its cost must be strictly less than the cost of any incorrect permutation:
\begin{equation}
    \mathcal{J}(I) < \mathcal{J}(S) \implies \| \tau \|_F < \| S W_{pre} - W_{pre} \|_F - \| \tau \|_F.
\end{equation}
This yields the strict inequality:
\begin{equation}
    2 \| \tau \|_F < \| S W_{pre} - W_{pre} \|_F.
\end{equation}
By definition, $\min_{S \neq I} \| S W_{pre} - W_{pre} \|_F = \delta_{min}$.
Therefore, if the task vector satisfies $2 \| \tau \|_F < \delta_{min}$, the cost of the correct assignment is strictly lower than any alternative.
The algorithm is mathematically forced to converge to $S^* = I$, which means $T^* P_{true} = I$, and therefore $T^* = P_{true}^{-1} = P_{true}^T$.
Applying this exact inverse to the protected model yields the fully recovered model:
\begin{equation}
    W^a = T^* W^p = P_{true}^T (P_{true} W_{ft}) = (P_{true}^T P_{true}) W_{ft} = I W_{ft} = W_{ft}.
\end{equation}
The recovery error is exactly $0$. 
This completes the proof. 

$\blacksquare$
Remark: This theorem exposes the fundamental fragility of permutation-based defenses (such as those applied to MLP blocks). 
In deep neural networks, the pre-trained weights $W_{pre}$ are highly diverse, meaning the row separation margin $\delta_{min}$ is significantly large. 
Conversely, the task vector $\tau$ inherently possesses a tiny magnitude. 
Because the condition $2 \| \tau \|_F < \delta_{min}$ is overwhelmingly satisfied in practical fine-tuning scenarios, AGA effortlessly bypasses the permutation defense with mathematically guaranteed perfect recovery.

\section{Related Work}

\subsection{Model Merging}

Model merging focuses on integrating multiple fine-tuned models into a unified architecture without incurring the substantial computational overhead of retraining. Foundational approaches, such as Task Arithmetic~\cite{ilharco2022editing}, achieve this by performing simple algebraic operations on task vectors. 
While subsequent methods like Ties-Merging~\cite{yadav2023ties} and DARE~\cite{yu2024language} attempt to refine this process through pruning and scaling, model merging strategies frequently struggle with severe parameter interference when fusing highly diverse tasks. 
To address these inherent conflicts and better preserve task-specific expertise, recent state-of-the-art paradigms have introduced advanced resolution mechanisms. 
Notably, CAT Merging~\cite{sun2025cat} and LOT Merging~\cite{wanglocalizing} have emerged as highly effective strategies that mitigate parameter conflicts and optimize the fusion trajectory. 
Given their superior merging utility and widespread adoption, we extensively employ Task Arithmetic, CAT Merging, and LOT Merging as the primary evaluation protocols to strictly assess both attack and defense mechanisms in this work.

\begin{figure}[t] 
    \centering
    \includegraphics[width=.8\linewidth]{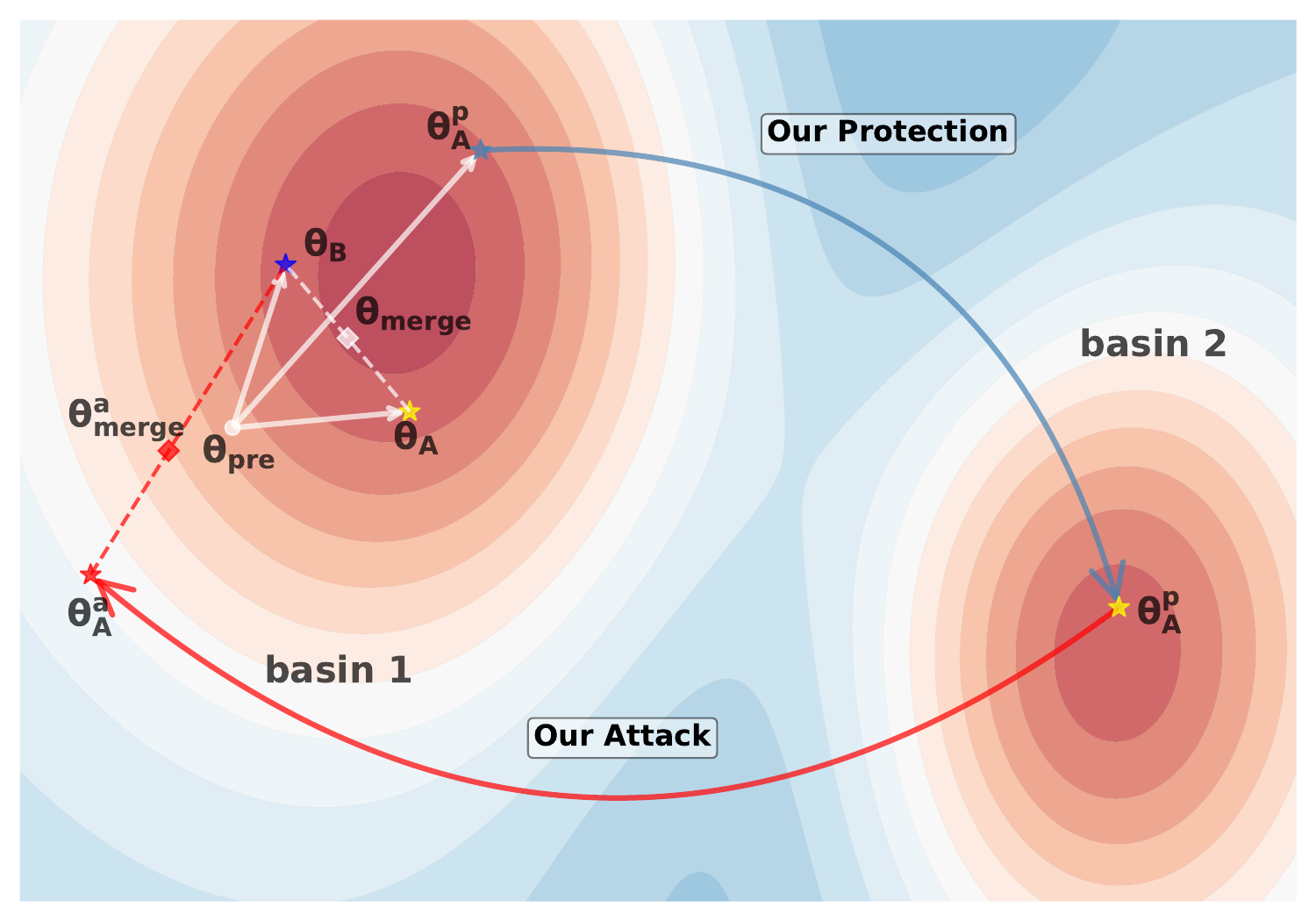}
    \caption{Loss landscape illustration of our protection ARF against our attack AGA.}
    \label{fig:loss_basin_protect_supp}
\end{figure}

\subsection{Proactive Protection in Model Merging}

The growing accessibility of model merging has raised critical security concerns regarding the unauthorized exploitation of proprietary model weights. 
Consequently, proactive protection methods have been rapidly developed to safeguard intellectual property~\cite{yu2024language, yu2025test}. 
These defenses generally fall into two categories: parameter-space obfuscation and structural alteration.
Params~\cite{junhao2025disrupting} and its advanced variant Params-P~\cite{junhao2025disrupting} introduce parameter-level safeguards by applying secret diagonal transformations to the fine-tuned weights, thereby disrupting the alignment required for unauthorized fusion.
MergeLock~\cite{wang2025model} elevates this concept by employing orthogonal matrices to strictly lock the parameter space. 
Furthermore, MergeBarrier~\cite{li2025not} extends protection beyond simple weight manipulation by structurally altering the topology of MLP modules via Taylor expansion, rendering standard task vector inversion mathematically ill-posed.
While these methods demonstrate empirical success against basic merging attempts, our work systematically exposes their shared geometric vulnerabilities.

\section{Loss Landscape Illustration of Our Protection}

To intuitively demonstrate how our defense neutralizes the Anchor-Guided Attack (AGA), we visualize the loss landscape geometry in Fig.~\ref{fig:loss_basin_protect_supp}. 
Fundamentally, our defense deliberately amplifies the distance between the fine-tuned model and the pretrained anchor, thereby bridging the magnitude disparity between the task vector and the pretrained weights.
Coupled with invertible linear transformations applied specifically to the Attention modules, ARF forcefully relocates the protected model $\theta_A^p$ from the optimal local minimum in basin 1 to an entirely distinct region in basin 2.
When AGA attempts to invert this protection using the pretrained anchor mathematically, the recovered model $\theta_A^a$ fails to return to the optimal center of basin 1, landing instead in a severely high-loss area. 
Consequently, when this sub-optimally recovered model is fused with another fine-tuned model $\theta_B$, the resulting merged model $\theta_{merge}^a$ is dragged into a high-loss boundary. 
This geometric visualization conclusively demonstrates that ARF disrupts AGA's inversion trajectory, significantly increasing loss and degrading the performance of any unauthorized merged models.

\begin{figure}[t]                       
  \centering                           
  \includegraphics[width=.6\linewidth]{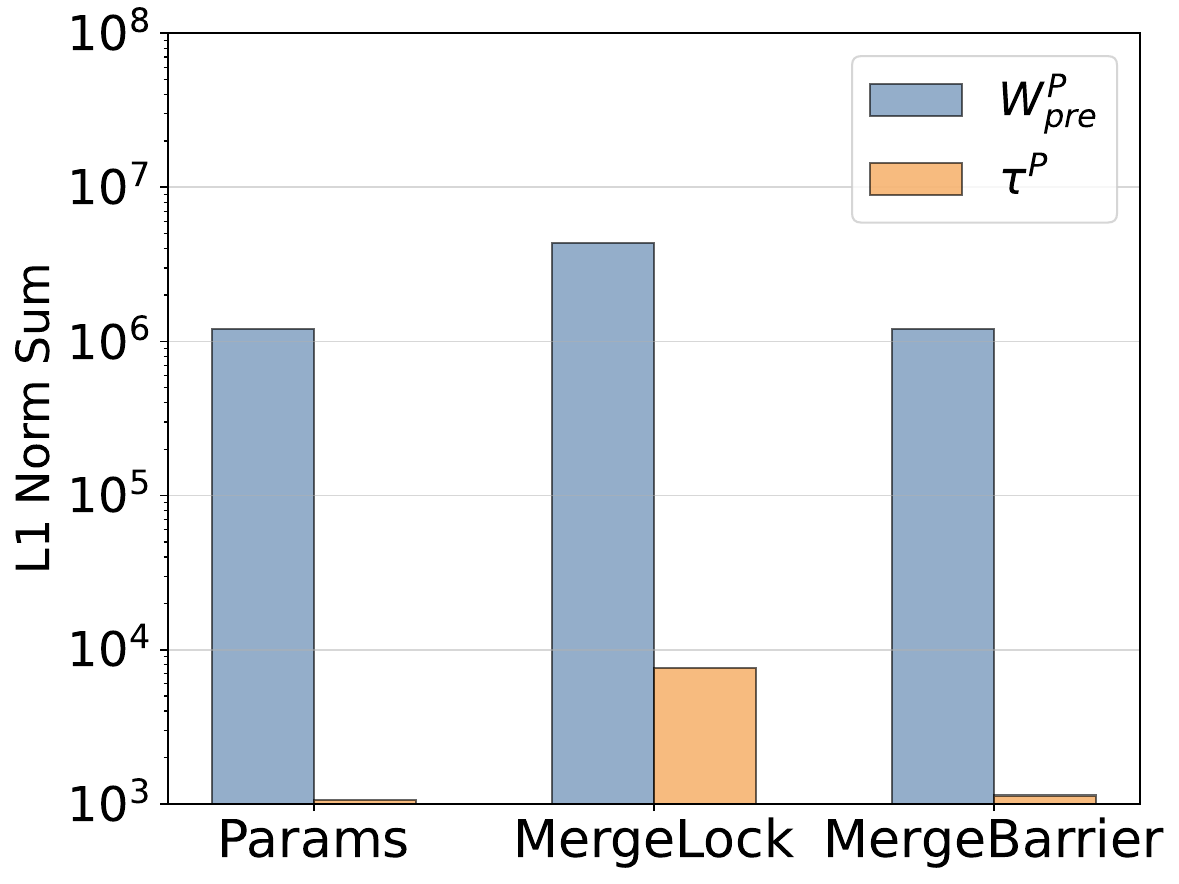}
  \caption{Comparison of the Frobenius norm between the protected pretrained weights ($W_{pre}^P$) and the protected task vector ($\tau^P$) on the ViT-L/14 model finetuned on Cars. 
  The visualization highlights a severe magnitude disparity, demonstrating that the protected pretrained anchor overwhelmingly dominates the task vector across all evaluated defenses.
           }
  \label{fig:L1_L14_supp}    
\end{figure}

\begin{figure}[t]                       
  \centering                           
  \includegraphics[width=.6\linewidth]{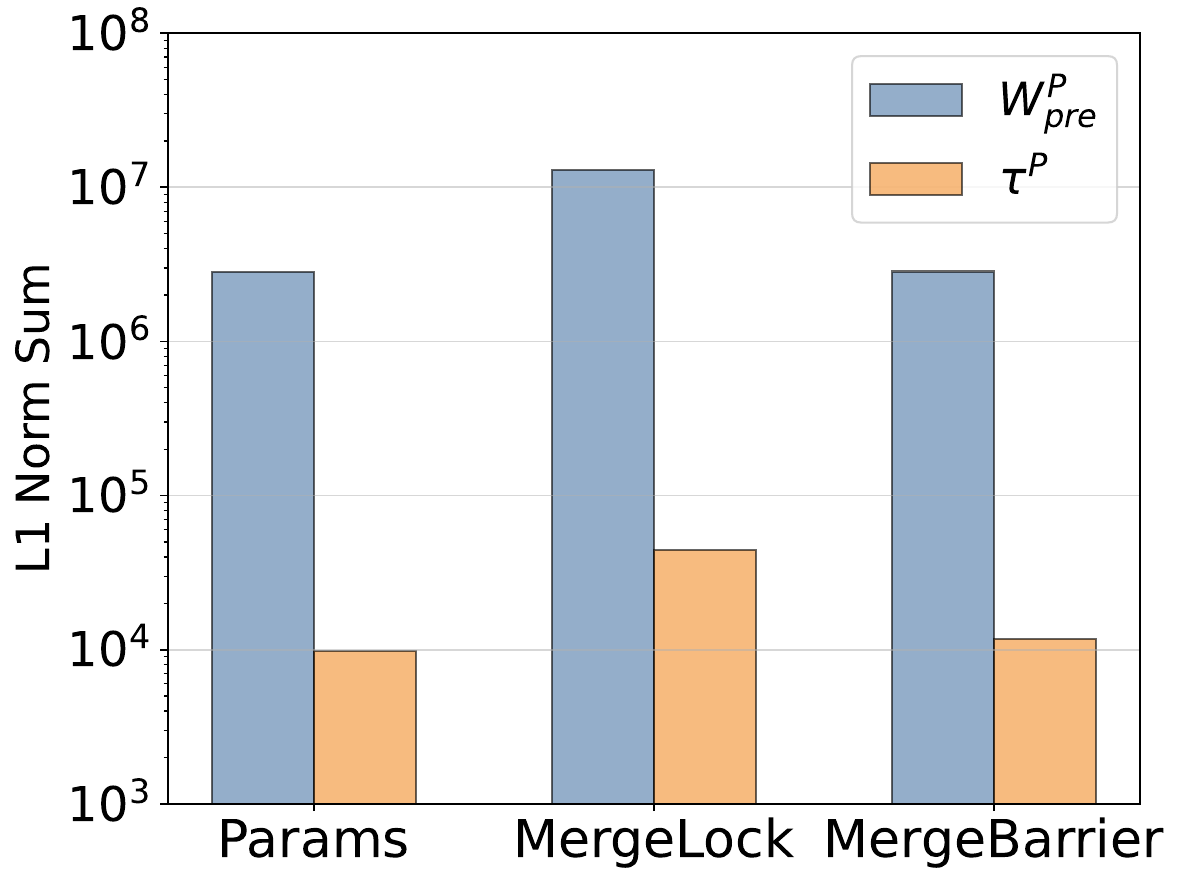}
  \caption{Comparison of the Frobenius norm between the protected pretrained weights ($W_{pre}^P$) and the protected task vector ($\tau^P$) on the GPT2 model finetuned on QQP. 
           }
  \label{fig:L1_GPT2_supp}    
\end{figure}

\begin{figure}[t]                       
  \centering                           
  \includegraphics[width=.6\linewidth]{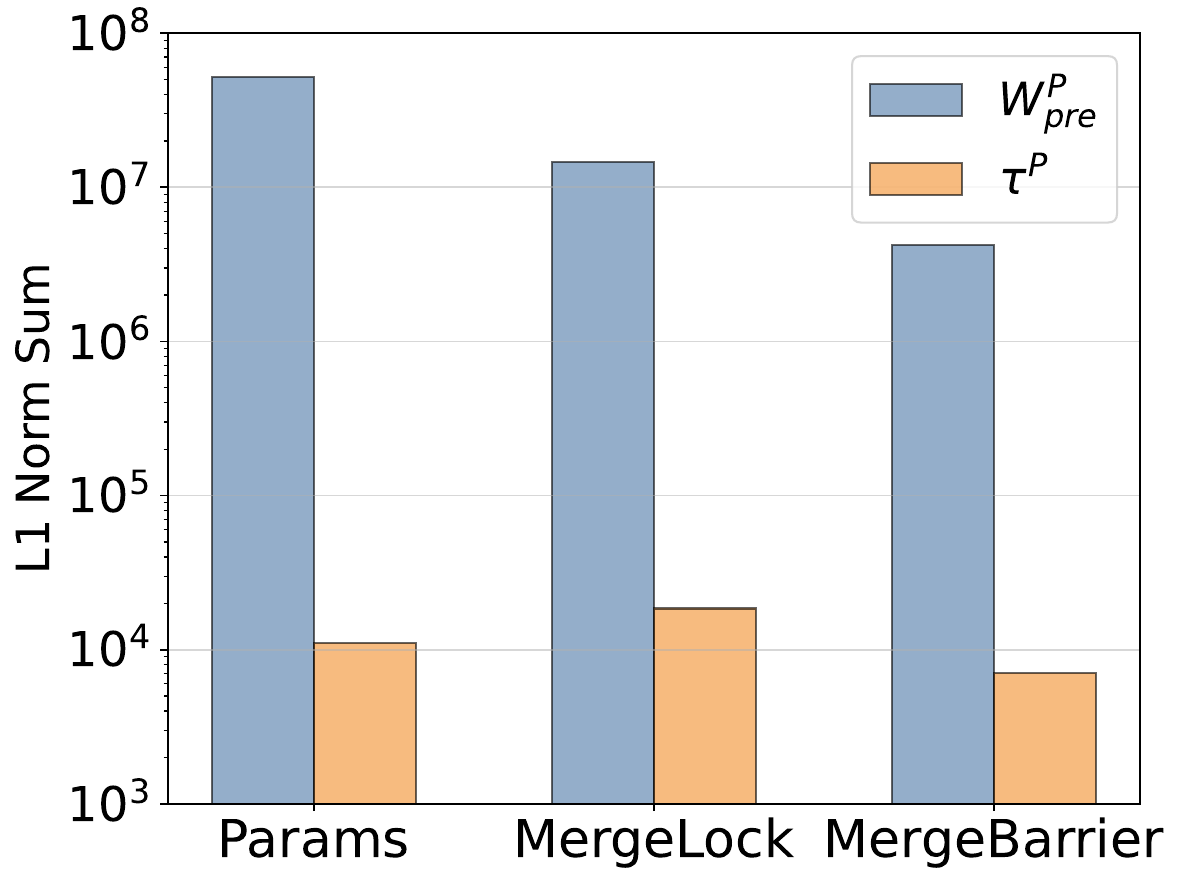}
  \caption{Comparison of the Frobenius norm between the protected pretrained weights ($W_{pre}^P$) and the protected task vector ($\tau^P$) on the Qwen2-7B model finetuned on Alpaca. 
           }
  \label{fig:L1_Qwen_3_supp}    
\end{figure}

\section{Analysis of Parameter Magnitude on More Backbones}

To further validate the systemic vulnerability identified in the main paper, we extend our empirical analysis of parameter magnitudes to a diverse set of larger and more complex architectures. 
As initially illustrated for ViT-B/32, the protected task vector $\tau^P$ exhibits a severe magnitude disparity when compared to the protected pretrained weights $W_{pre}^P$.
The corresponding Frobenius norm comparisons are explicitly detailed in Fig.~\ref{fig:L1_L14_supp} for ViT-L/14, Fig.~\ref{fig:L1_GPT2_supp} for GPT-2, and Fig.~\ref{fig:L1_Qwen_3_supp} for the Qwen2-7B architecture. 
Across all evaluated vision and language backbones, we consistently observe that the magnitude of $W_{pre}^P$ is substantially larger than that of $\tau^P$, typically by two to three orders of magnitude. 
This overwhelming dominance confirms that the geometric vulnerability exploited by our Anchor-Guided Attack (AGA) is not an isolated phenomenon. 
Instead, it is a fundamental and universal flaw inherent to current linear protection paradigms, perfectly explaining AGA's high success rate across various model scales.

\section{More Experimental Results}
\subsection{Extended Evaluation of Our Attack on ViT-L/14 and GPT-2}

To rigorously substantiate the cross-architecture and cross-modal generalization capabilities of our Anchor-Guided Attack (AGA), we provide extended evaluations on the GPT-2 backbone.
As detailed in Table~\ref{tab:GPT2_supp}, AGA poses a formidable threat to Natural Language Processing (NLP) models. 
On the GPT-2 architecture, the unprotected baseline achieves an average score of 68.79\% across standard NLP benchmarks. 
When parameter-level defenses such as Params are applied, the performance drops significantly to 51.44\%. 
AGA effortlessly circumvents this obfuscation, reconstructing the weights to achieve an impressive 67.22\%. 
Even against structure-altering defenses like MergeBarrier, AGA elevates the accuracy to a highly competitive 59.86\%. 
These comprehensive supplementary results conclusively prove that the systemic magnitude disparity exploited by AGA represents a universal vulnerability, posing a critical security risk across diverse model scales and modalities.

\begin{table}[t]
    \centering
    \caption{
    Evaluation of protected-task performance under various defense mechanisms and our AGA framework using the GPT-2 backbone. 
    {\color{Red} {$\uparrow \Delta$ }} presents the accuracy recovered by AGA relative to the only-protected setting.
    } 
    \label{tab:GPT2_supp}
    
    \setlength{\tabcolsep}{1.2pt} 
    \renewcommand{\arraystretch}{1.5} 

    \newcommand{\upR}[1]{\,\makebox[0pt][l]{\raisebox{-0ex}{\scalebox{1}{(\textcolor{red}{$\uparrow$ #1})}}}}
    \newcommand{\headerDelta}{\,\makebox[0pt][l]{\raisebox{-0ex}{\scalebox{1}{({\color{Red} {$\uparrow \Delta$ }})}}}}

    \resizebox{0.9\textwidth}{!}{%
        \begin{NiceTabular}{
            l 
            c 
            *{7}{>{\centering\arraybackslash}p{1.3cm}} 
            c@{\hskip 35pt}
        }
            \hline
            Protect & Attack & CoLA & MNLI & MRPC & QNLI & QQP & RTE & SST-2 & Avg\headerDelta \\
            \hline
            
            \Block[fill=white]{1-1}{-} & - & 68.26 & 68.30 & 71.81 & 66.90 & 80.79 & 46.21 & 79.24 & 68.79 \\
            \hline
            
            \Block[fill=white]{2-1}{Params~\cite{junhao2025disrupting}} & - & 66.03 & 34.42 & 41.18 & 54.99 & 65.16 & 42.57 & 55.73 & 51.44 \\[-1.4ex]
             & AGA & 68.26 & 62.87 & 71.81 & 65.50 & 78.53 & 45.85 & 77.75 & 67.22\upR{15.78} \\
            \hline
            
            \Block[fill=white]{2-1}{Params-P~\cite{junhao2025disrupting}} & - & 65.29 & 32.08 & 41.24 & 54.26 & 64.72 & 42.10 & 54.19 & 50.55\\[-1.4ex]
             & AGA & 68.17 & 62.06 & 71.69 & 64.31 & 77.26 & 45.66 & 76.51 & 66.52\upR{15.97} \\
            \hline
            
            \Block[fill=white]{2-1}{MergeLock~\cite{wang2025model}} & - & 63.85 & 37.04 & 64.46 & 51.36 & 57.88 & 41.29 & 67.75 & 54.80 \\[-1.4ex]
             & AGA & 68.07 & 64.00 & 71.81 & 63.35 & 80.45 & 46.21 & 77.75 & 67.38\upR{12.58} \\
            \hline
            
            \Block[fill=white]{2-1}{MergeBarrier~\cite{li2025not}} & - & 60.97 & 34.89 & 42.89 & 57.24 & 59.45 & 42.38 & 56.99 & 50.69 \\[-1.4ex]
             & AGA & 65.19 & 56.18 & 68.14 & 60.26 & 70.40 & 46.21 & 52.64 & 59.86\upR{9.17} \\
            \hline
             
        \end{NiceTabular}
    }
\end{table}

\subsection{Extended Evaluation of Our Protection on ViT-L/14 and GPT-2}

To corroborate the robust defensive capabilities of Anchor-Repulsive Fine-tuning (ARF) presented in the main manuscript, we extend our evaluation to the GPT-2 backbone. 
According to Table~\ref{tab:GPT-2_supp}, ARF successfully lowers the merged model's utility to 52.27\%, outperforming MergeGuard's 58.20\%. 
When attacked by AGA, the ARF-protected GPT-2 model maintains a suppressed accuracy of 54.91\%, continuing to provide a stronger protective barrier than the unattacked baseline defense. 
These consistent supplementary results thoroughly validate that ARF's localized optimization objective effectively neutralizes parameter-space vulnerabilities across diverse model scales and modalities.

\begin{table}[htbp]
    \centering
    \caption{Evaluation of protected-task performance under protection and AGA using the GPT-2 backbone.}
    \label{tab:GPT-2_supp}

    \setlength{\tabcolsep}{1.2pt} 
    \renewcommand{\arraystretch}{1.5}

    \resizebox{0.85\textwidth}{!}{%
        \begin{NiceTabular}{
            l 
            c 
            *{7}{>{\centering\arraybackslash}p{1.3cm}} 
            c
        }
            \hline
            Protect & Attack & CoLA & MNLI & MRPC & QNLI & QQP & RTE & SST-2 & Avg \\
            \hline
            
            \Block[fill=white]{1-1}{-} & - & 68.26 & 68.30 & 71.81 & 66.90 & 80.79 & 46.21 & 79.24 & 68.79 \\
            \hline
            
            \Block[fill=white]{1-1}{MergeGuard~\cite{chen2025defending}} & \Block[fill=white]{2-1}{-} & 65.31 & 48.23 & 65.06 & 55.27 & 64.41 & 42.80 & 66.35 & 58.20 \\[-1.4ex]
            \Block[fill=white]{1-1}{ARF} &  & 61.09 & 35.29 & 60.55 & 51.97 & 52.61 & 41.35 & 63.06 & 52.27 \\
            \hline
            
            \Block[fill=white]{1-1}{ARF} & AGA & 63.50 & 37.78 & 62.29 & 53.62 & 60.25 & 42.00 & 64.94 & 54.91\\
            \hline
             
        \end{NiceTabular}%
    }
\end{table}

\subsection{Extended Evaluation of Our Protection on CAT Merging and LOT Merging}

To explicitly demonstrate the robust generalization of our Anchor-Repulsive Fine-tuning (ARF) across different parameter fusion strategies, we extend our defensive evaluation to include CAT Merging~\cite{sun2025cat} and LOT Merging~\cite{wanglocalizing} on the ViT-B/32 backbone. 
As detailed in Table~\ref{tab:CAT_protect_supp} and Table~\ref{tab:LOT_protect_supp}, ARF consistently neutralizes unauthorized merging utilities regardless of the specific merging algorithm. 
In the absence of attacks, ARF suppresses the average merged accuracy to near-random levels, achieving a mere 5.63\% under CAT Merging and 4.70\% under LOT Merging.
These results significantly outperform the baseline MergeGuard, which only reduces the performance to 37.18\% and 44.16\%, respectively.

Crucially, ARF maintains its formidable protective barrier even when subjected to our aggressive Anchor-Guided Attack (AGA). 
Under CAT Merging, the accuracy of the ARF-protected model under attack is strictly restricted to 28.17\%. 
Similarly, under LOT Merging, the attacked ARF model yields an accuracy of 34.28\%. 
In both scenarios, the performance of our defense under active attack remains substantially lower than that of the unattacked MergeGuard baseline. 
This compelling empirical evidence confirms that ARF successfully eliminates exploitable parameter-space vulnerabilities across various merging strategies, establishing it as a highly generalized and merging-agnostic defense framework.

\begin{table}[t]
    \centering
    \caption{
    Evaluation of protected-task performance under protection and AGA using the ViT-B/32 backbone. Models are merged via CAT Merging~\cite{sun2025cat}.
    }
    \label{tab:CAT_protect_supp}
    
    \setlength{\tabcolsep}{1.2pt} 
    \renewcommand{\arraystretch}{1.5} 

    \resizebox{\textwidth}{!}{%
        \begin{NiceTabular}{
            l 
            c 
            *{8}{>{\centering\arraybackslash}p{1.4cm}} 
            c
        }
            \hline
            Protect & Attack & SUN397 & Cars & RESISC45 & EuroSAT & SVHN & GTSRB & MNIST & DTD & Avg \\
            \hline
            
            \Block[fill=white]{1-1}{-} & - & 65.80 & 60.42 & 69.47 & 80.38 & 83.03 & 59.47 & 98.47 & 50.43 & 70.93 \\
            \hline
            
            \Block[fill=white]{1-1}{MergeGuard~\cite{chen2025defending}} & \Block[fill=white]{2-1}{-} & 45.02 & 31.19 & 33.86 & 48.43 & 42.11 & 24.86 & 35.92 & 36.01 & 37.18 \\[-1.4ex]
            \Block[fill=white]{1-1}{ARF} &  & 0.30  & 0.35  & 2.04  & 12.38 & 12.11 & 5.37  & 10.12 & 2.33  & 5.63  \\
            \hline
             \Block[fill=white]{1-1}{ARF} & AGA & 38.78 & 25.06 & 21.22 & 28.81 & 38.58 & 18.04 & 29.75 & 25.11 & 28.17 \\
            \hline
        \end{NiceTabular}%
    }
\end{table}

\begin{table}[t]
    \centering
    \caption{Evaluation of protected-task performance under protection and AGA using the ViT-B/32 backbone. Models are merged via LOT Merging~\cite{wanglocalizing}}
    \label{tab:LOT_protect_supp}
    
    \setlength{\tabcolsep}{1.2pt} 
    \renewcommand{\arraystretch}{1.5}

    \resizebox{\textwidth}{!}{%
        \begin{NiceTabular}{
            l 
            c 
            *{8}{>{\centering\arraybackslash}p{1.4cm}} 
            c
        }
            \hline
            Protect & Attack & SUN397 & Cars & RESISC45 & EuroSAT & SVHN & GTSRB & MNIST & DTD & Avg \\
            \hline
            
            \Block[fill=white]{1-1}{-} & - & 66.96 & 65.70 & 75.65 & 92.80 & 92.65 & 82.98 & 98.85 & 57.66 & 79.16 \\
            \hline
            
            \Block[fill=white]{1-1}{MergeGuard~\cite{chen2025defending}} & \Block[fill=white]{2-1}{-} & 49.29 & 35.16 & 48.66 & 60.82 & 48.90 & 33.58 & 36.07 & 40.83 & 44.16 \\[-1.4ex]
            \Block[fill=white]{1-1}{ARF} &  & 0.22  & 0.57  & 3.91  & 12.58 & 6.09  & 2.46  & 10.28 & 1.49  & 4.70  \\
            \hline
            
             \Block[fill=white]{1-1}{ARF} & AGA & 40.17 & 29.04 & 28.10 & 48.25 & 41.76 & 26.35 & 31.04 & 29.51 & 34.28 \\
            \hline
        \end{NiceTabular}%
    }
\end{table}

\begin{table}[t]
    \centering
    \caption{Standalone performance of individual models fine-tuned with and without protection methods with ViT-L/14 as the backbone.}
    \label{tab:L14_finetune_supp}
    
    \resizebox{\textwidth}{!}{%
        \begin{tabular}{lccccccccc}
            \toprule
            Method & ~SUN397~ & ~~Cars~~ & RESISC45 & ~EuroSAT & ~~SVHN~~ & ~~GTSRB~ & ~~MNIST~ & ~~~DTD~~ & ~~~Avg~~ \\
            \midrule
            Individual  & 82.23 & 92.35 & 98.86 & 99.86 & 98.11 & 99.24 & 99.69 & 84.15 & 94.31 \\
            MergeGuard~\cite{chen2025defending} & 73.74 & 83.70 & 91.53 & 97.72 & 95.90 & 98.31 & 99.36 & 81.22 & 90.19 \\
            ARF                & 81.01 & 91.28 & 98.17 & 99.02 & 97.83 & 99.16 & 99.63 & 83.41 & \textbf{93.69} \\
            \bottomrule
        \end{tabular}%
    }
\end{table}

\begin{table}[t]
    \centering
    \caption{
    Standalone performance of individual models fine-tuned with and without protection methods with GPT-2 as the backbone.}
    \label{tab:GPT2_finetune_supp}
    
    \resizebox{0.8\textwidth}{!}{%
        \begin{tabular}{lcccccccc}
            \toprule
            Method & ~CoLA~ & ~~MNLI~~ & ~MRPC~ & ~QNLI~ & ~QQP~ & ~RTE~ & ~SST-2~ & ~~~Avg~~ \\
            \midrule
            Individual  & 76.80 & 81.99 & 80.39 & 88.27 & 89.64 & 65.34 & 91.17 & 81.94 \\
            MergeGuard~\cite{chen2025defending} & 73.83 & 76.15 & 77.46 & 83.59 & 85.80 & 62.39 & 84.66 & 77.70 \\
            ARF                & 75.02 & 81.12 & 79.76 & 86.96 & 89.04 & 64.51 & 90.23 & \textbf{80.95} \\
            \bottomrule
        \end{tabular}
    }
\end{table}

\subsection{Extended Evaluation of Standalone Performance on ViT-L/14 and GPT-2}

To further validate that Anchor-Repulsive Fine-tuning (ARF) preserves the original task-specific expertise without causing detrimental performance degradation, we extend our evaluation of standalone model utility to the ViT-L/14 and GPT-2 architectures. 
As detailed in Table~\ref{tab:L14_finetune_supp}, ARF achieves a remarkable average accuracy of 93.69\% on the larger ViT-L/14 backbone. 
This result closely approaches the 94.31\% performance of the unprotected individual models, noticeably outperforming the baseline defense, MergeGuard, which suffers a significant drop to 90.19\% due to its global constraints.

This high-fidelity preservation translates consistently to the Natural Language Processing domain.
According to Table~\ref{tab:GPT2_finetune_supp} for the GPT-2 backbone, ARF maintains a strong average score of 80.95\%. 
In stark contrast to MergeGuard, which degrades the original utility down to 77.70\%, ARF incurs a negligible drop of less than 1\% compared to the 81.94\% unprotected baseline. 
These supplementary findings conclusively demonstrate that by restricting the repulsive force exclusively to the attention modules, ARF successfully secures the models against unauthorized merging while keeping the standalone performance strictly intact across diverse scales and modalities.

\subsection{Standalone Performance of Attacked Models}
To comprehensively evaluate the fidelity of our Anchor-Guided Attack (AGA), we further analyze the standalone classification accuracy of individual models post-attack. 
As detailed in Table~\ref{tab:standalone_attack_supp}, AGA proves to be an entirely non-destructive inversion process. 
For the ViT-B/32 backbone, the original unprotected fine-tuned model achieves an average accuracy of 88.81\%. 
When AGA is applied directly to this unprotected model, the recovered performance remains exactly at 88.81\%. 
More importantly, when attacking models protected by state-of-the-art defenses like Params and MergeLock, AGA consistently restores the standalone accuracy to approximately 88.80\%. 
Even against the structure-altering MergeBarrier, the attacked model retains a highly competitive 88.01\%. These empirical results conclusively demonstrate that our attack framework effectively dismantles protection mechanisms while perfectly preserving the task-specific expertise of the original models.

\begin{table}[ht]
    \centering
    \caption{Standalone classification accuracy of individual attacked models using the ViT-B/32 backbone.}
    \label{tab:standalone_attack_supp}
    
    \renewcommand{\arraystretch}{1.4} 
    
    \resizebox{\textwidth}{!}{
        \begin{NiceTabular}{>{\raggedright\arraybackslash}p{2.6cm}>{\centering\arraybackslash}p{1.4cm} *{9}{>{\centering\arraybackslash}p{1.5cm}}}
            \hline
            Protect & Attack & SUN397 & Cars & RESISC45 & EuroSAT & SVHN & GTSRB & MNIST & DTD & Avg \\
            \hline
            
            {-} & \Block[fill=white]{1-1}{-} & 74.54 & 76.40 & 91.67 & 97.89 & 97.39 & 98.94 & 99.65 & 73.99 & 88.81 \\
            \hline
            {-} & \Block[fill=white]{1-1}{AGA} & 74.78 & 76.46 & 91.59 & 97.98 & 97.25 & 98.96 & 99.62 & 73.84 & 88.81 \\[-1.4ex]
            
            {Params~\cite{junhao2025disrupting}} & \Block[fill=white]{1-1}{AGA} & 74.78 & 76.46 & 91.57 & 97.98 & 97.22 & 98.95 & 99.62 & 73.78 & 88.80 \\[-1.4ex]
            
            {Params-P~\cite{junhao2025disrupting}} & \Block[fill=white]{1-1}{AGA} & 74.86 & 76.88 & 91.37 & 98.08 & 97.33 & 99.01 & 99.66 & 73.88 & 88.88 \\[-1.4ex]
            
            {MergeLock~\cite{wang2025model}} & \Block[fill=white]{1-1}{AGA} & 74.79 & 76.45 & 91.56 & 97.98 & 97.23 & 98.95 & 99.62 & 73.83 & 88.80 \\[-1.4ex]
            
            {MergeBarrier~\cite{li2025not}} & \Block[fill=white]{1-1}{AGA} & 74.68 & 76.46 & 91.50 & 97.94 & 91.17 & 98.94 & 99.63 & 73.73 & 88.01 \\
            \hline
        \end{NiceTabular}
    }
\end{table}

\subsection{Ablation Studies}
We empirically evaluated the sensitivity of $\rho$ and $\lambda_{dist}$ on ViT-B/32, measuring both the average finetuned accuracy and the average protected accuracy of our AFR protection under AGA attack.
The results in Table~\ref{Hyperparameter sensitivity} demonstrate that ARF is highly robust to hyperparameter variations.

\begin{table}[h]
    \centering
    \caption{Hyperparameter sensitivity in ARF with ViT-B/32.}
    \label{Hyperparameter sensitivity}
    \renewcommand{\arraystretch}{1.2} 
    \resizebox{.85\textwidth}{!}{%
    \begin{NiceTabular}{>{\raggedright\arraybackslash}p{2.6cm}>{\centering\arraybackslash}p{1.4cm} *{6}{>{\centering\arraybackslash}p{1.4cm}}}
    \toprule
     $\rho (\lambda_{dist}=1.0)$ & 0.02 & 0.03 & 0.04 & 0.05 &  0.06  &  0.07 & 0.08 \\ \midrule
    {Finetuend} & 88.32  & 88.27 & 88.16 & 88.11 & 88.05 & 88.02 & 87.94 \\
    {Protected } & 28.75  & 28.01 & 27.80 &  27.42 & 27.17 & 26.89 & 26.36  \\
    \bottomrule
     $\lambda_{dist} (\rho =0.05)$ & 0.7 & 0.8 & 0.9 & 1.0 &  1.1  &  1.2 & 1.3 \\ \midrule
    {Finetuend} & 88.38  & 88.30 & 88.25 & 88.11 & 87.89 & 87.53 & 87.01 \\
    {Protected } & 29.26  & 28.73 & 27.84 & 27.42 & 27.06 & 26.51 & 26.17  \\
    \bottomrule
        \end{NiceTabular}
    }
\end{table}

\end{document}